\documentclass{article}

    \usepackage[final]{neurips_2020}

\usepackage[utf8]{inputenc} 
\usepackage[T1]{fontenc}    
\usepackage{hyperref}       
\usepackage{url}            
\usepackage{booktabs}       
\usepackage{amsfonts}       
\usepackage{nicefrac}       
\usepackage{microtype}      
\usepackage{epsfig}
\usepackage{graphicx}
\usepackage{subfigure}

\def\eg{\emph{e.g.~}}
\def\etal{{\em et al.~}}
\def\ie{\emph{i.e.~}}

\title{A reanalysis and correction to ObjectNet}
\title{ObjectNet Dataset: Reanalysis and Correction}

\author{
Ali Borji \\
\texttt{aliborji@gmail.com} 
\thanks{Code and data is available at: \url{https://github.com/aliborji/ObjectNetReanalysis.git}}
}

\begin{document}

\maketitle

\begin{abstract}

Recently, Barbu \etal introduced a dataset called ObjectNet which includes objects in daily life situations. They showed a dramatic performance drop of the state of the art object recognition models on this dataset. Due to the importance and implications of their results regarding generalization ability of deep models, we take a second look at their findings. We highlight a major problem with their work which is applying object recognizers to the scenes containing multiple objects rather than isolated objects. The latter results in around 20-30\% performance gain using our code. Compared with the results reported in the ObjectNet paper, we observe that around 10-15\% of the performance loss can be recovered, without any test time data augmentation. In accordance with Barbu {\em et al.}'s conclusions, however, we also conclude that deep models suffer drastically on this dataset. Thus, we believe that ObjectNet remains a challenging dataset for testing the generalization power of models beyond datasets on which they have been trained.

\end{abstract}

\section{Introduction}

Object recognition is arguably the most important problem at the heart of computer vision. Application of an already known 
convolutional neural network architecture (CNN) known as LeNet~\cite{lecun1998gradient}, albeit with new tips and tricks~\cite{krizhevsky2012imagenet}, revolutionized not only computer vision but also several other areas including machine learning, natural language processing, time series prediction, etc. With the initial excitement gradually damping, researchers have started to study the shortcomings of deep learning models and question their generalization power. From prior research (\eg~\cite{azulay2019deep,recht2019imagenet,goodfellow2014explaining}), we already know that CNNs: a) often fail when applied to transformed versions of the same object. In other words, they are not invariant to transformations such as translation\footnote{CNNs are not invariant to translation but are equivariant to it.}, in-plane and in-depth rotation, scale, lighting, and occlusion, b) lack out of distribution generalization. Even after being exposed to so many different instances of the same object category they are not good at learning that concept. In stark contrast, humans are able to generalize only from few examples, and c) are sensitive to small image perturbations (\ie adversarial examples~\cite{goodfellow2014explaining}).



Several datasets have been proposed in the past to train and test deep models and to study their generalization ability (\eg ImageNet~\cite{imagenet}, CIFAR~\cite{krizhevsky2009learning}, NORB~\cite{lecun2004learning}, iLab20M~\cite{borji2016ilab}). In a recent effort, Barbu~\etal introduced a dataset called ObjectNet which has less bias than other datasets\footnote{This dataset, however, has it own biases. It consists of indoor objects that are available to many people, are mobile, are not too large, too small, fragile or dangerous.} and is supposed to be used solely as a test set\footnote{Unlike some benchmarks (\eg ImageNet) that hide their test images, images in ObjectNet dataset will be publicly available. See~\url{http://objectnet.dev}.}. Objects in this dataset are pictured by Mechanical Turk workers using a mobile app in a variety of backgrounds, rotations, and imaging viewpoints. It contains 50,000 images from 313 categories, out of which 113 are in common with the ImageNet, and comes with a licence that disallows the researchers to finetune models on it. Barbu~\etal find that the state of the art object detectors\footnote{They mean object recognizers!} perform drastically lower than their corresponding performance on the ImageNet dataset (about 40-45\% drop). Here, we revisit the Barbu~{\em et al.}'s study and seek to answer how much performance of models drops on this dataset compared with their performance on ImageNet. Due to this, here we limit our analysis to the 113 overlapped categories.



\begin{figure}[t]
    \centering
    \includegraphics[width=\textwidth]{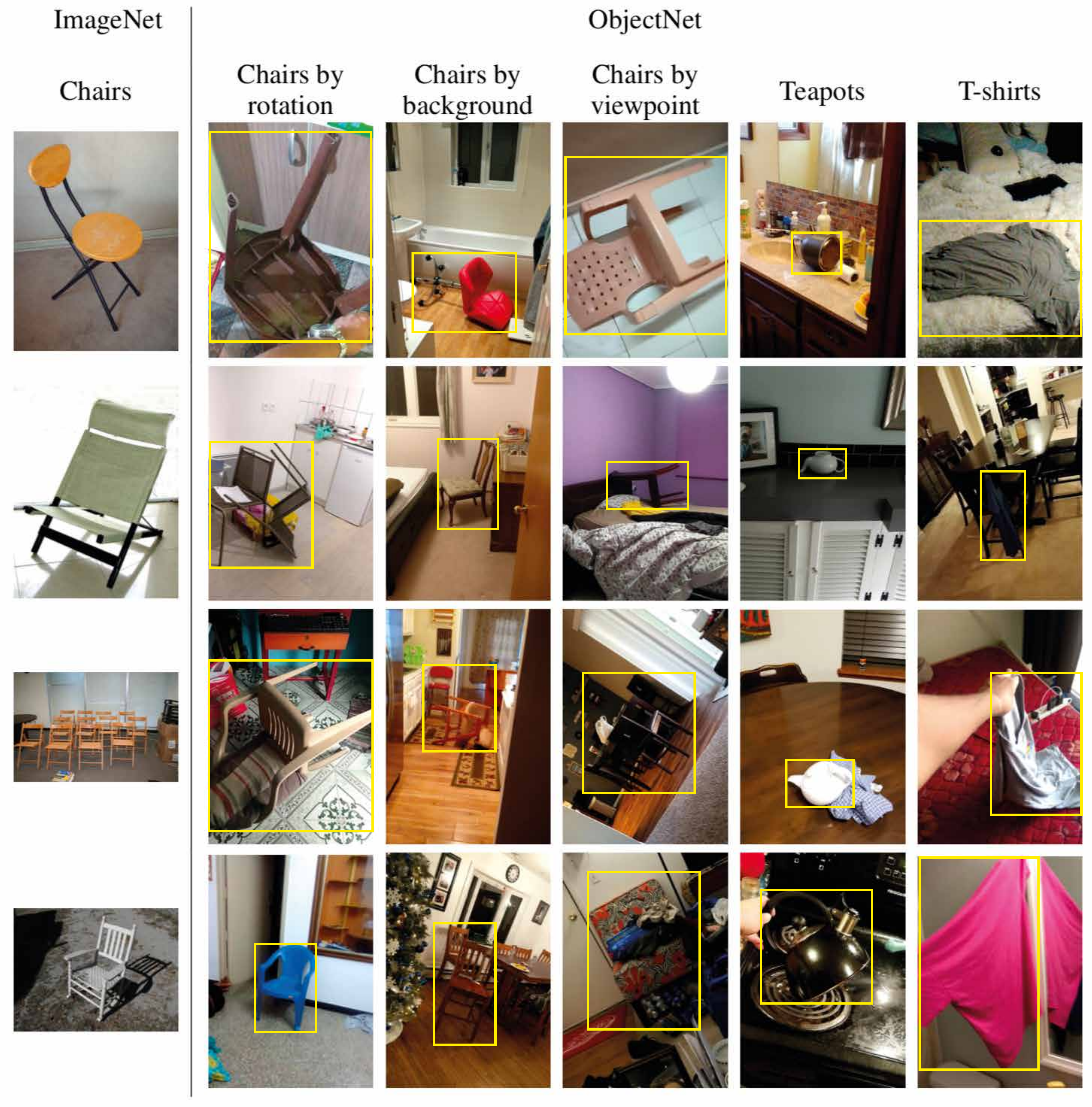}
    \caption{Sample images from the ObjectNet dataset from chairs, teapots and t-shirts categories, along with their corresponding object bounding boxes. The left panel shows objects in the ImageNet dataset. As it can be seen ImageNet scenes often contain a single isolated object. There are also many images in the ImageNet dataset with multiple objects, but we speculate that scenes in the ObjectNet dataset have more number of images on average~\cite{imagenet}. This figure is a modified version of the Figure 2 in~\cite{barbu2019objectnet}.}
    \label{fig:box}
    \vspace{-10pt}
\end{figure}


Barbu~{\em et al.}'s work is a great contribution to the field to answer how well object recognition models generalize to the real world situations. It, however, suffers from a major flaw which is making no distinction between "object recognition" and "object detection". They use the term "object detector" to refer to object recognition models. This bring along several concerns:
\begin{itemize}
    \item Instead of applying object recognition models to individual object bounding boxes, they apply them to cluttered scenes containing multiple objects. Alternatively, they could have run object detection models on scenes and measured detection performance (\ie mean average precision; mAP).
    
    \item Object detection and object recognition are two distinct, but related, tasks. Each has its own models, datasets, and evaluation measures. For example, as shown in Fig.~\ref{fig:box}, images in object recognition datasets (\eg ImageNet~\cite{imagenet}) often contain a single object, usually from a closeup view, whereas scenes in object detection datasets (\eg MS COCO~\cite{lin2014microsoft}) have multiple objects. Due to this, object characteristics in the two types of datasets might be different. For example, objects are often smaller in detection datasets compared to recognition datasets~\cite{singh2018analysis}. 
    
    \item It is true that ImageNet also includes a fair amount of images that have more than one object, but it appears that ObjectNet images are more cluttered and have higher number of objects, although we have not quantified this since this dataset does not come with object-level labels. This discussion is also related to the difference between "scene understanding" and "object recognition". To understand a complex scene, we look around, fixate on individual objects to recognize 
    them, and accumulate information over fixations to perform more complex tasks such as answering a question or describing an event.
    
    \item One might argue that the top-5 metric somewhat takes care of scenes with multiple objects. While this might be true to some degree, it certainly does not entail that it is better to train or test object recognition models on scenes with multiple objects. One problem is labeling such images. For example, a street scene with a car, a pedestrian, trees and buildings is given only one label which adds noise to the data. Another problem is invariance. Given that individual objects can undergo many transformations, feeding scenes with multiple objects to models exacerbates the problem and leads to combinatorially much more variations which are harder to learn. A better approach would be focusing on individual objects. As humans, we also move our eyes around and place our fovea on one object at a time to recognize it, using information from our peripheral vision. The analogue of this task still does not exist in computer vision. This task is reminiscent of object detection but there are subtle differences which will be elaborated on later in the discussion section.
    
\end{itemize}

 Here, we first annotate the objects in the ObjectNet scenes and then apply a number of deep object recognition models to only object bounding boxes.

\section{Experiments and Results}

We employ six deep models including AlexNet~\cite{krizhevsky2012imagenet}, VGG-19~\cite{simonyan2014very}, GoogLeNet~\cite{szegedy2015going}, ResNet-152~\cite{resnet} 
Inception-v3~\cite{szegedy2016rethinking}, and MNASNet~\cite{tan2019mnasnet}. AlexNet, VGG-19, and GoogLeNet have also been employed in the ObjectNet paper~\cite{barbu2019objectnet}. We use the pytorch implementation of these models\footnote{https://pytorch.org/docs/stable/torchvision/models.html}. Notice that the code from the ObjectNet paper
is still not available. Due to possible inconsistency in our code and the code in ObjectNet as well as different data processing methods, in addition to bounding boxes, we also run the models on the entire scenes. This allows us to study whether and how much performance varies across these two conditions. 


\subsection{Bounding box annotation}

\begin{figure}
\centering
\subfigure{\hspace{-100pt}
\includegraphics[width=1.5\linewidth]{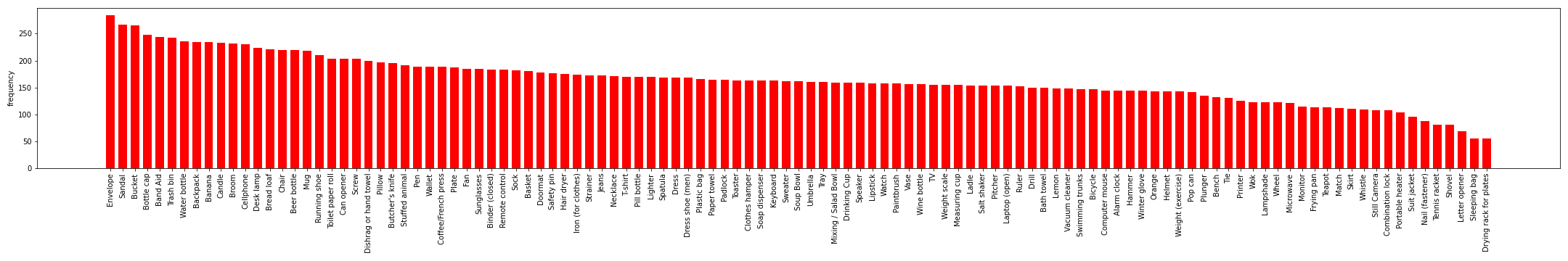}}
\vspace{-10pt}
\caption{Frequency of the images per category over the 113 categories of the ObjectNet dataset overlapped with the ImageNet.}
\label{fig:freq}
\end{figure}

The 113 categories of the ObjectNet dataset, overlapped with the ImageNet, contain 18,574 images in total. On this subset, the average number of images per category is 164.4 (min = 55, max = 284).  Fig.~\ref{fig:freq} shows the distribution of number of images per category on this dataset. We drew\footnote{The annotation tools include: \url{https://github.com/aliborji/objectAnnotation} and \url{https://github.com/tzutalin/labelImg}.} a bounding box around the object corresponding to the category label of each image. If there were multiple objects from that category we tried to included all of them in the bounding box (\eg chairs around a table). Some example scenes and their corresponding bounding boxes are given in Fig.~\ref{fig:box}.

\subsection{Results}

\begin{figure}
    \centering
    \includegraphics[width=1\textwidth]{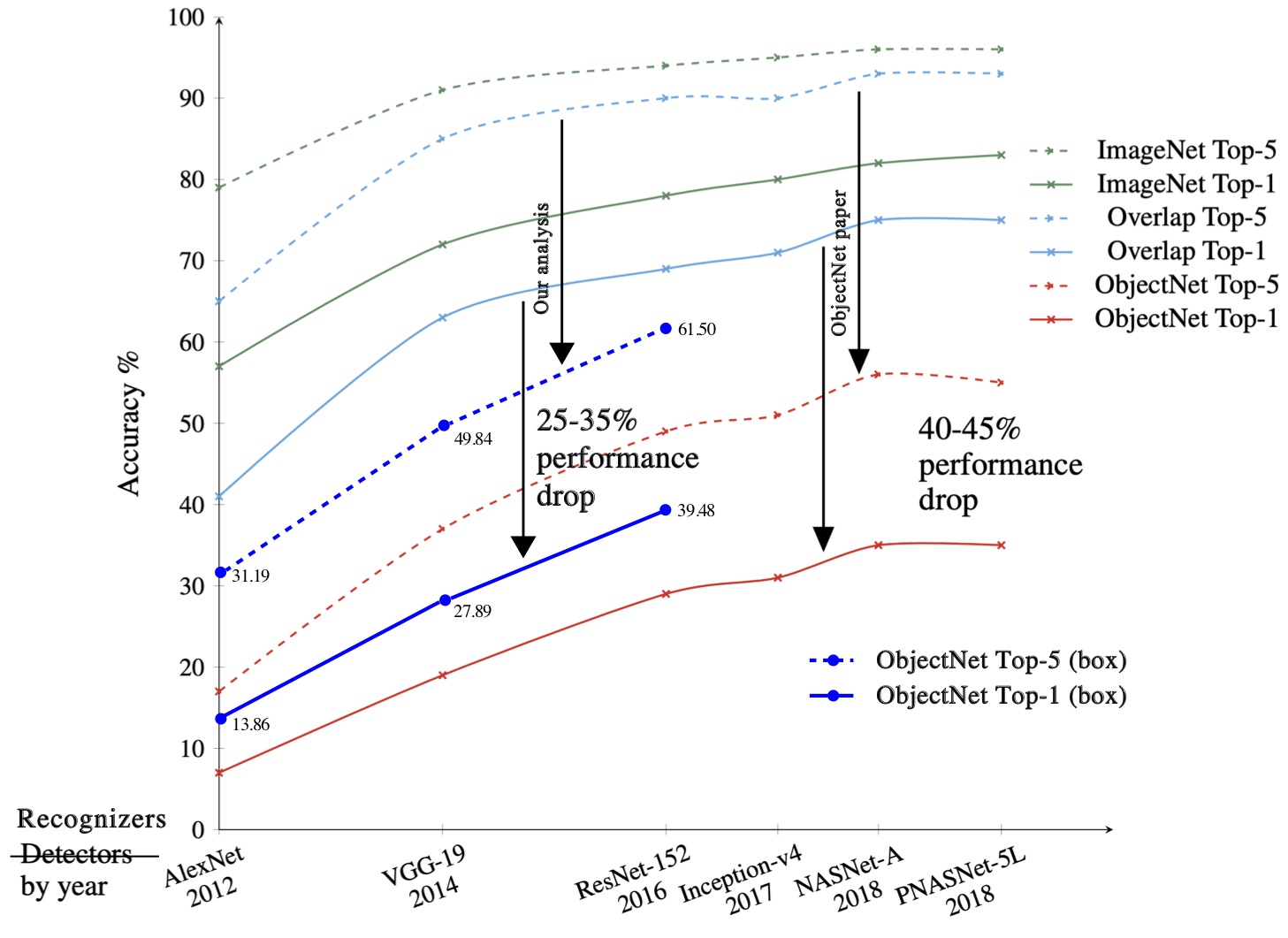}
    \caption{Performance of the state of the art object recognition models over the ObjectNet dataset. Results of our analysis by applying AlexNet~\cite{krizhevsky2012imagenet}, VGG-19~\cite{simonyan2014very}, and ResNet-152~\cite{resnet} on bounding boxes are shown in blue. As it can be seen, applying models to boxes instead of the entire scene improves the performance about 10\%, but sill much lower than results over the ImageNet dataset (overlap cases). \textbf{Note: This is not the best possible score on this dataset since we are not performing test time data augmentation. Using the original code in the ObjectNet paper with bounding boxes will likely improve this results.} This figure is a modified version of the Figure 1 in~\cite{barbu2019objectnet}.}
    \label{fig:ObjNet}
\end{figure}

Fig.~\ref{fig:ObjNet} shows an overlay of our results on the same figure from the ObjectNet paper. As it can be seen, applying models to boxes instead of the entire scene improves the performance about 10\%, but sill much lower than the results over the ImageNet dataset. The gap, however, is narrower now.

\begin{figure}
    \centering
    \includegraphics[width=.8\textwidth]{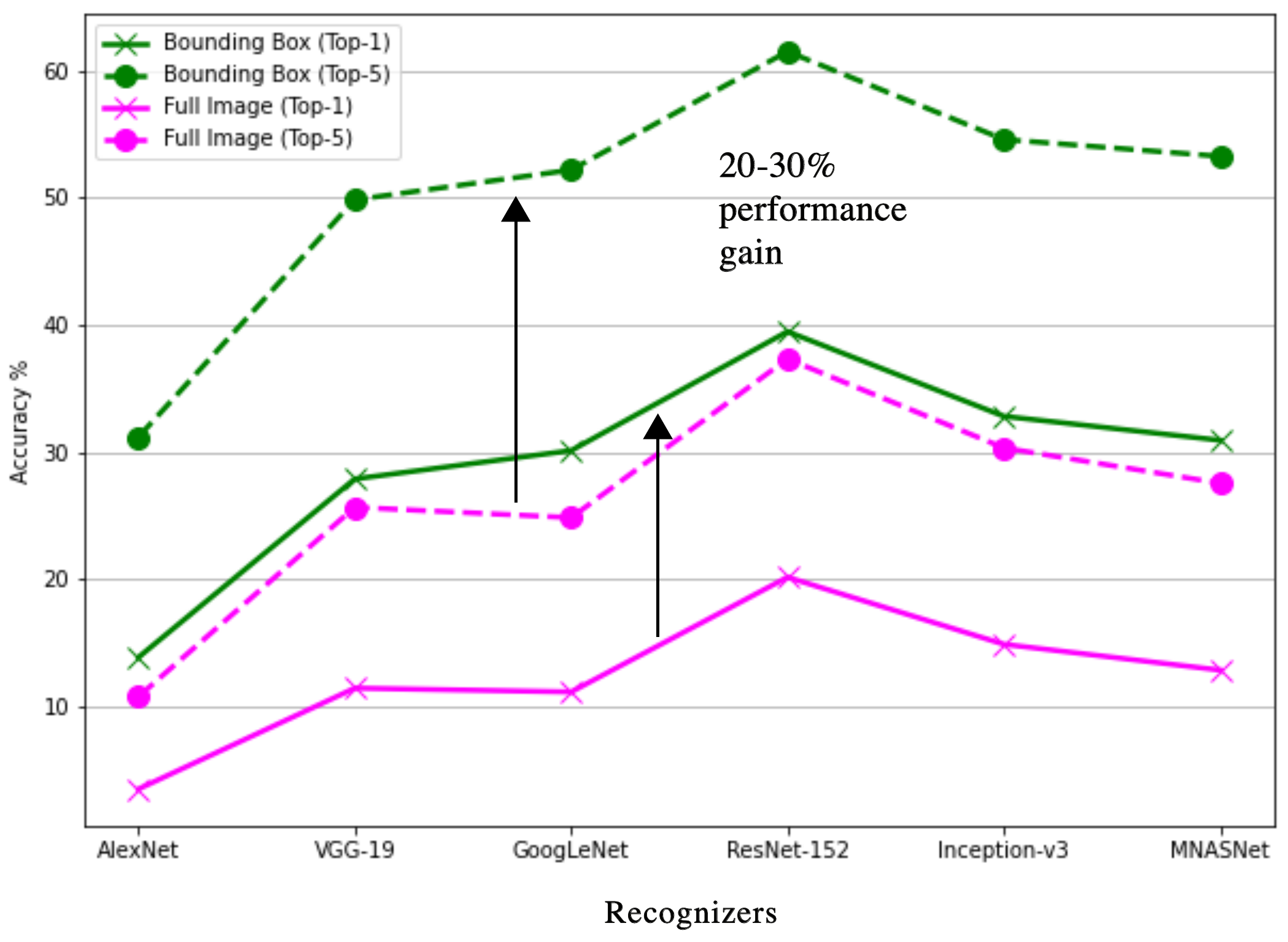}
    \caption{Results of our analysis by running six deep models on bounding boxes and whole images over the ObjectNet dataset (Corresponding to the overlap case in Fig.~\ref{fig:ObjNet}). Models perform significantly better on boxes than full images. }
    \label{fig:Ours}

\end{figure}

Since the code of~\cite{barbu2019objectnet} is not available, we could not run the exact pipeline used by them on the bounding boxes. It is possible that they might have performed a different data normalization or test time data augmentation\footnote{Such as rotation, scale, color jittering, cropping, etc.} to achieve better results. To remedy this, we also applied models to the whole image to study how much performance varies. Results are shown in Fig.~\ref{fig:Ours}. We find that:
\begin{enumerate}
    \item Focusing on a small image region containing only a single object increases the performance significantly by around 20-30\% across all tested models. 
    \item Our results on the whole scene are lower than Barbu~{\em et al.}'s results (which are also on the whole scene). This entails that applying their code to bounding boxes will likely improve the performance even more than our results using boxes. Assuming 25\% gain in performance on top of their best results, when using boxes, will still not close the gap in performance. Please see Fig.~\ref{fig:ObjNet}. We will discuss this further in the next section.
\end{enumerate}


\begin{figure}
\centering
\subfigure{\hspace{-100pt}
\includegraphics[width=1.5\linewidth]{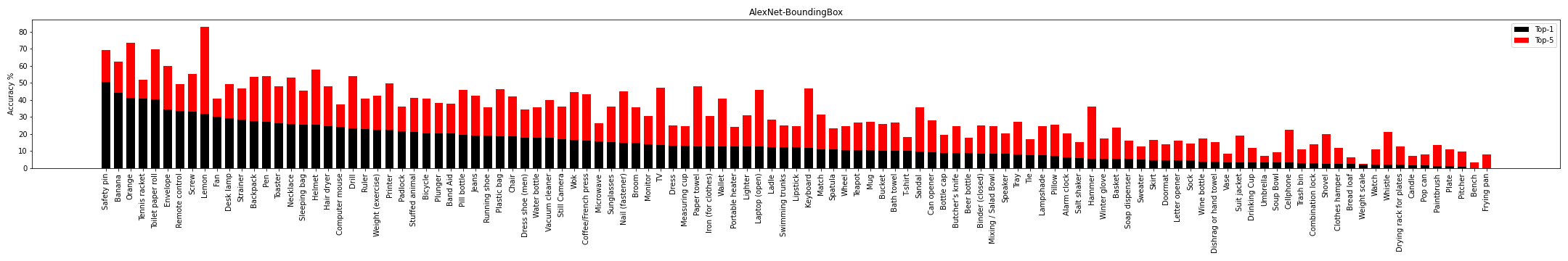}}
\vspace{-10pt}
\hspace{-100pt}

\subfigure{\hspace{-100pt}
\includegraphics[width=1.5\linewidth]{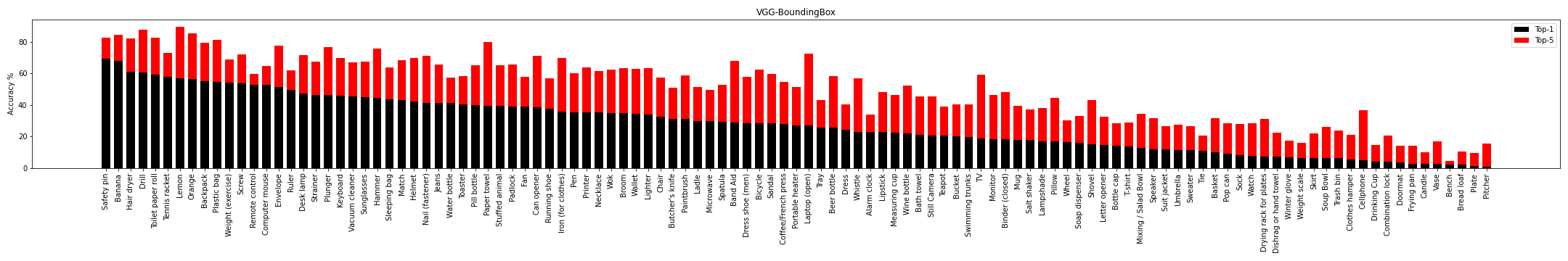}}
\vspace{-10pt}
\hspace{-100pt}

\subfigure{\hspace{-100pt}
\includegraphics[width=1.5\linewidth]{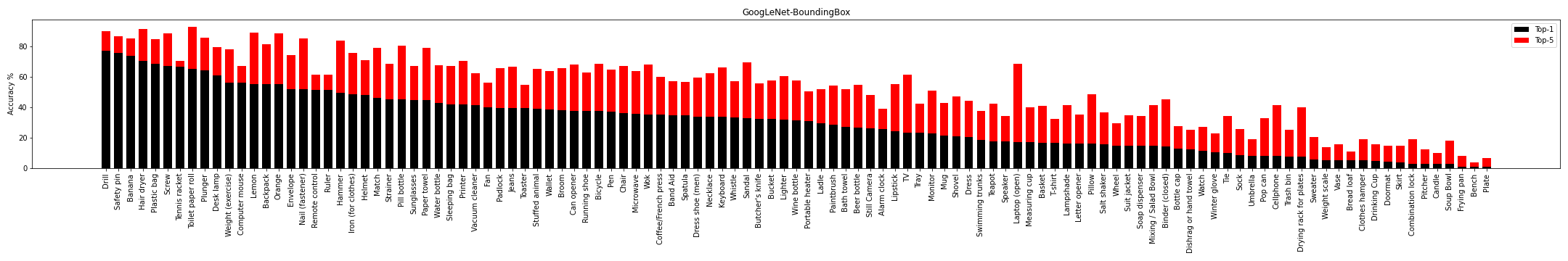}}
\vspace{-10pt}
\hspace{-100pt}

\subfigure{\hspace{-100pt}
\includegraphics[width=1.5\linewidth]{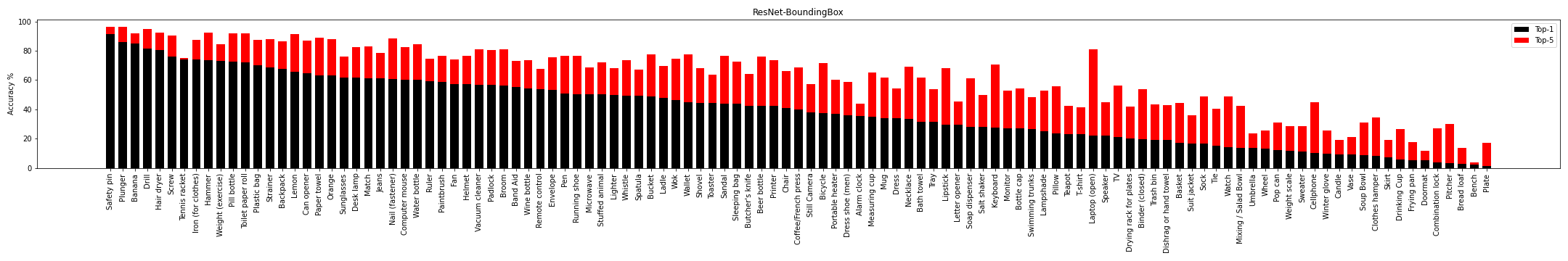}}
\vspace{-10pt}
\hspace{-100pt}

\subfigure{\hspace{-100pt}
\includegraphics[width=1.5\linewidth]{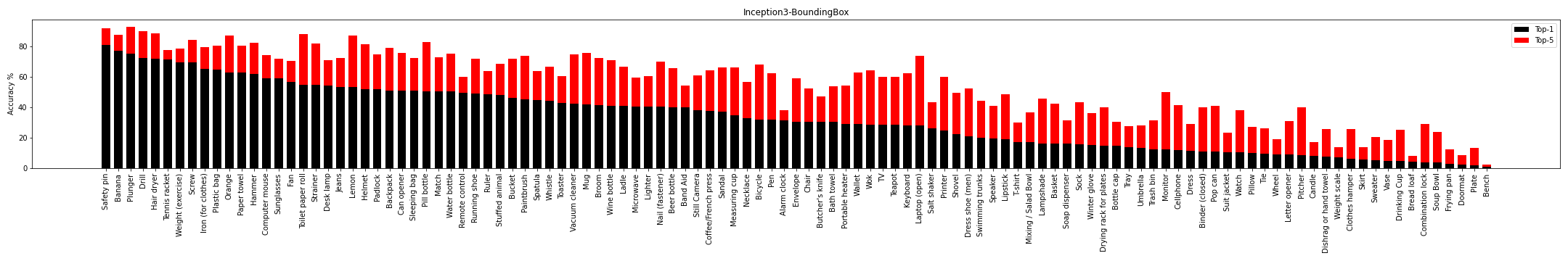}}
\vspace{-10pt}
\hspace{-100pt}

\subfigure{\hspace{-100pt}
\includegraphics[width=1.5\linewidth]{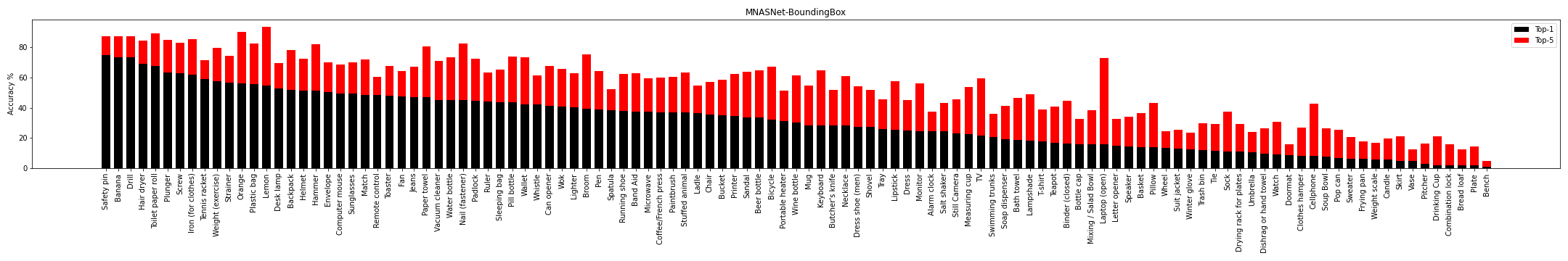}}
\vspace{-15pt}
\hspace{-100pt}

\caption{Performance of models on object bounding boxes}
\label{fig:boxResult}
\end{figure}

\begin{figure}
\centering
\subfigure{\hspace{-100pt}
\includegraphics[width=1.5\linewidth]{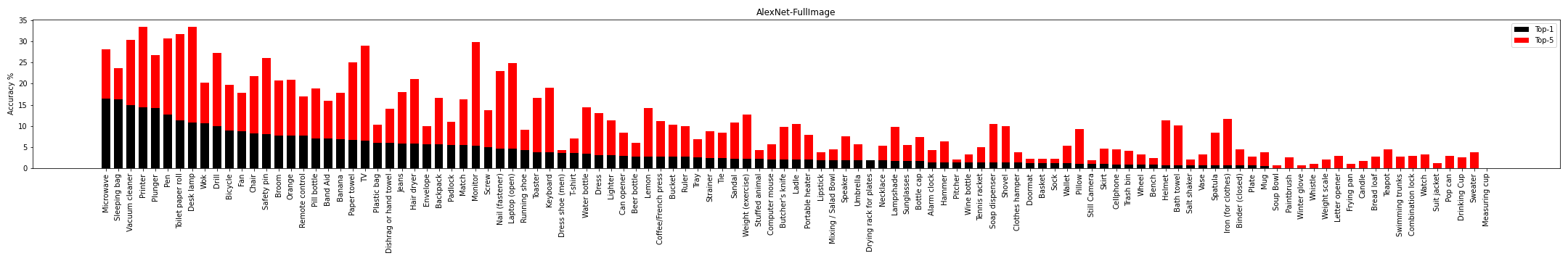}}
\vspace{-10pt}
\hspace{-100pt}

\subfigure{\hspace{-100pt}
\includegraphics[width=1.5\linewidth]{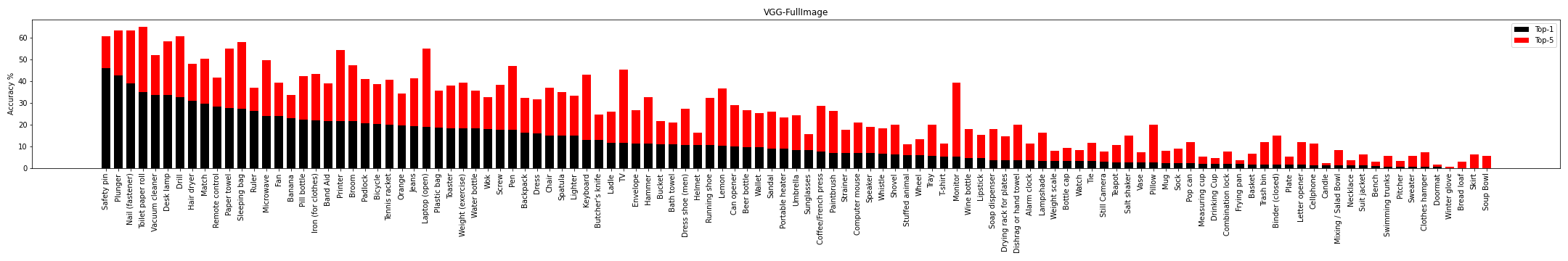}}
\vspace{-10pt}
\hspace{-100pt}

\subfigure{\hspace{-100pt}
\includegraphics[width=1.5\linewidth]{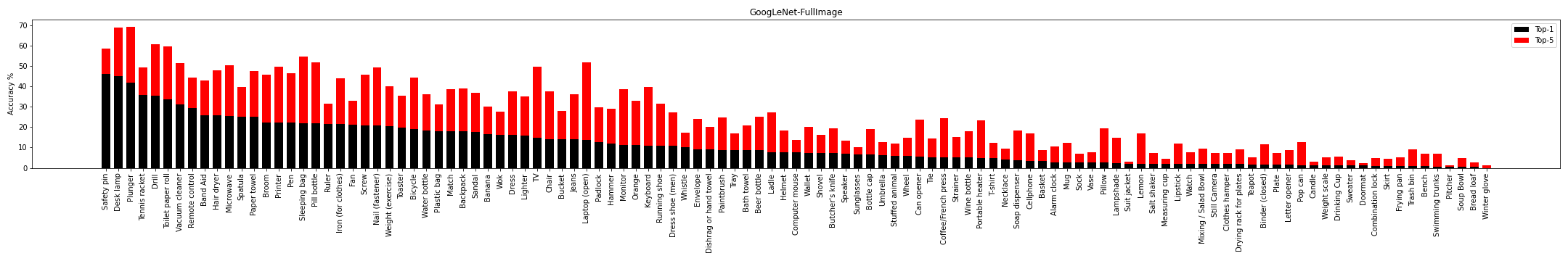}}
\vspace{-10pt}
\hspace{-100pt}

\subfigure{\hspace{-100pt}
\includegraphics[width=1.5\linewidth]{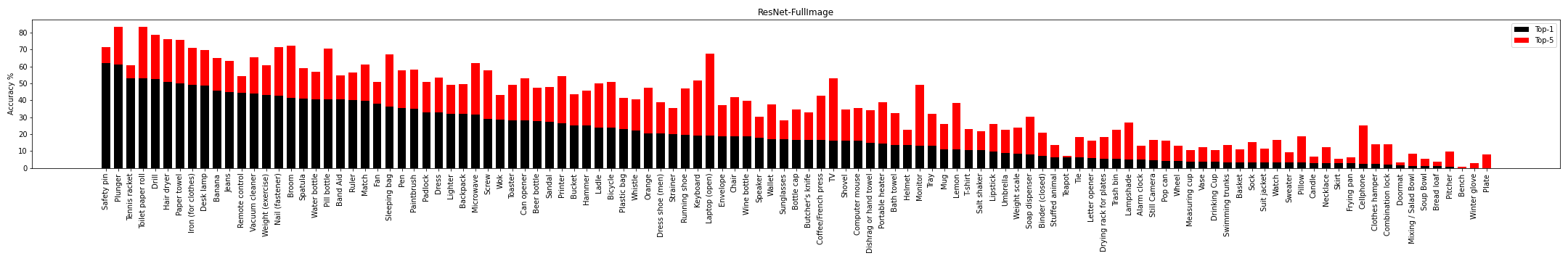}}
\vspace{-10pt}
\hspace{-100pt}

\subfigure{\hspace{-100pt}
\includegraphics[width=1.5\linewidth]{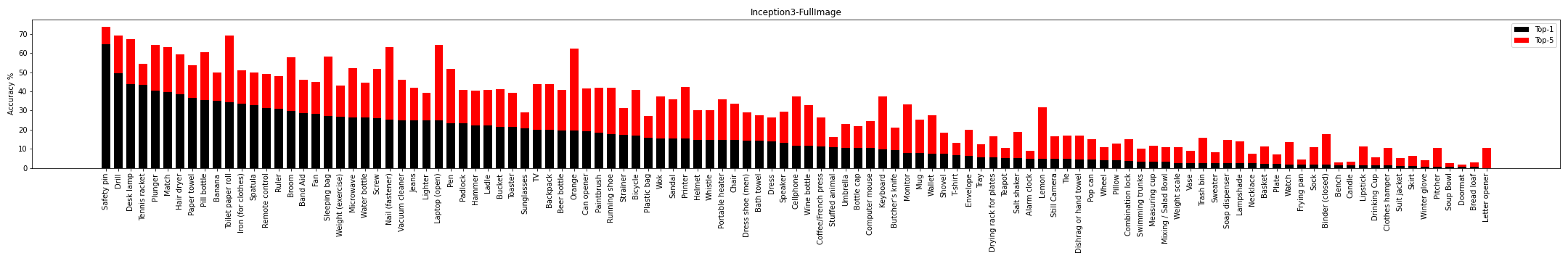}}
\vspace{-10pt}
\hspace{-100pt}

\subfigure{\hspace{-100pt}
\includegraphics[width=1.5\linewidth]{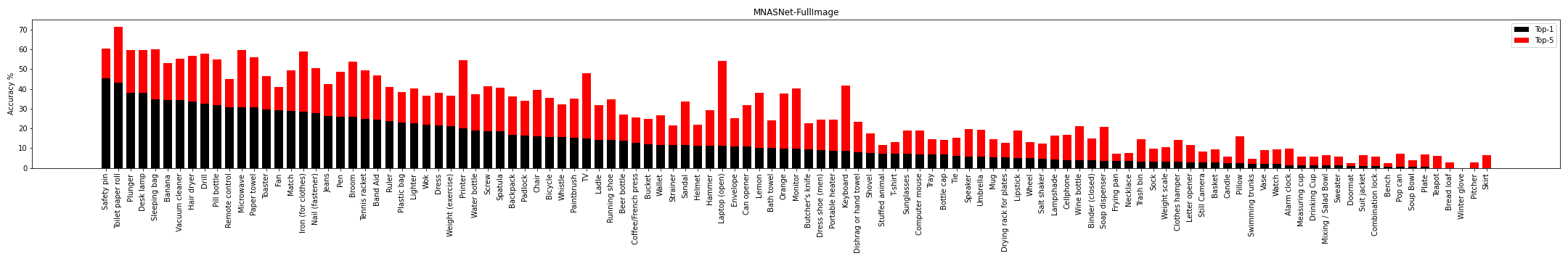}}
\vspace{-15pt}
\hspace{-100pt}

\caption{Performance of models over the entire image (full image)}
\label{fig:fullResult}
\end{figure}

Break down of performance over 113 categories for each of the 6 tested models is shown in Fig.~\ref{fig:boxResult} (over isolated objects) and Fig.~\ref{fig:fullResult} (over full image). Interestingly, in both cases, almost all models (except the GoogLeNet on isolated objects and the AlexNet on full image) perform the best over the \emph{safety pin} category. Inspecting the images from this class, we found that they have a single safety pin often hold by a person. The same story is true about the \emph{banana} class which is the second easy category using the bounding boxes. This object becomes much harder to recognize when using the full image (26.88\% vs. 70.3\% using boxes) which highlights the benefit of applying models to isolated objects rather than scenes.



\begin{figure}
    \hspace{-50pt}
    \includegraphics[width=1.2\textwidth]{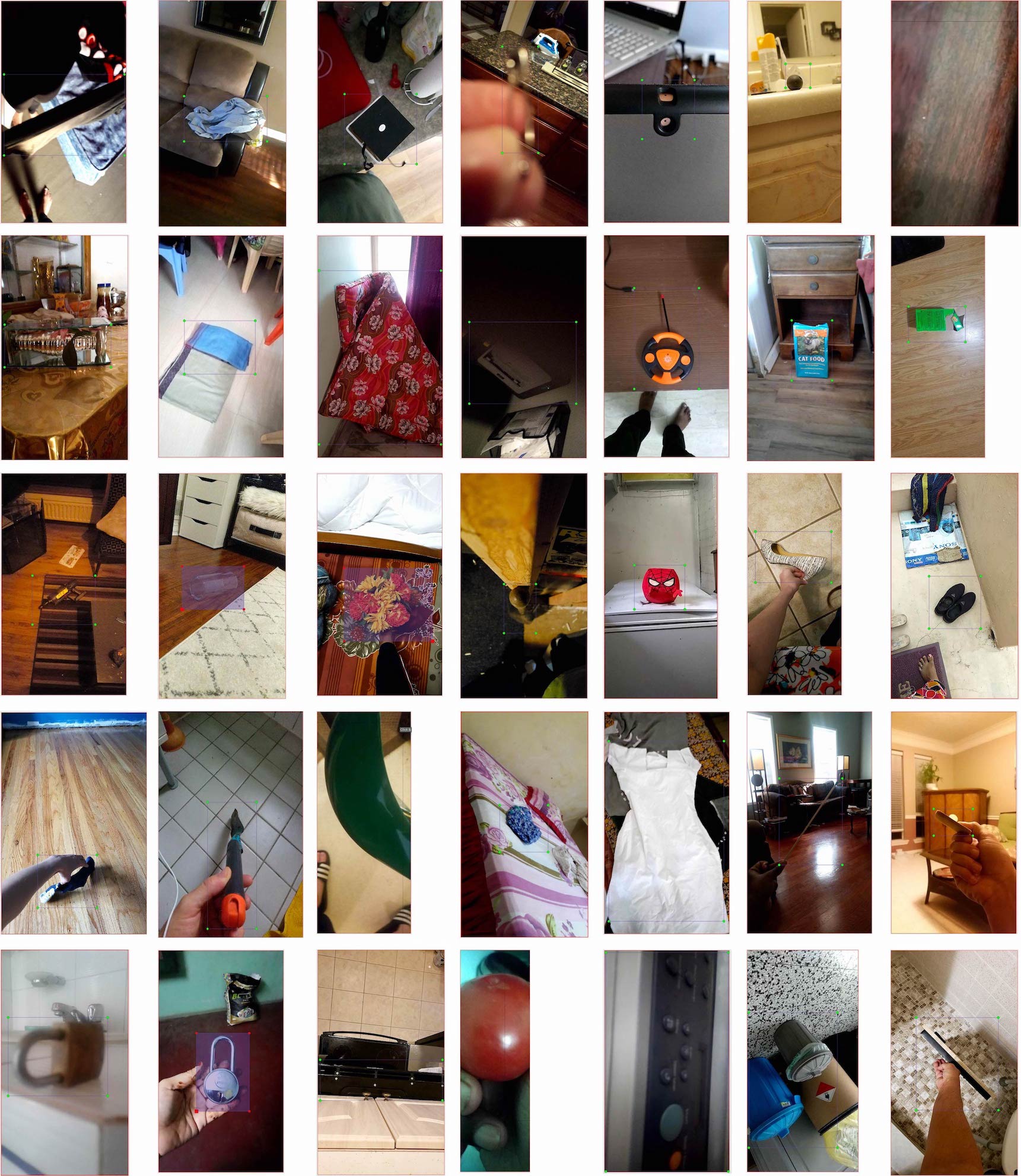}
    \caption{A selection of challenging objects that are hard to be recognized by humans. Can you guess the category of the annotated objects in these images? Keys are as follows:  \newline
    row 1: \emph{(skirt, skirt, desk lamp, safety pin, still camera, spatula, tray)},  \newline 
    row 2: \emph{(vase, pillow, sleeping bag, printer, remote control, pet food container, detergent)},  \newline
    row 3: \emph{(vacuum cleaner, vase, vase, shovel, stuffed animal, sandal, sandal)},  \newline
    row 4: \emph{(sock, shovel, shovel, skirt, skirt, match, spatula)},  \newline
    row 5: \emph{(padlock, padlock, microwave, orange, printer, trash bin, tray)}
    }
    \label{fig:Samples1}
\end{figure}

\begin{figure}
    \hspace{-50pt}
    \includegraphics[width=1.2\textwidth]{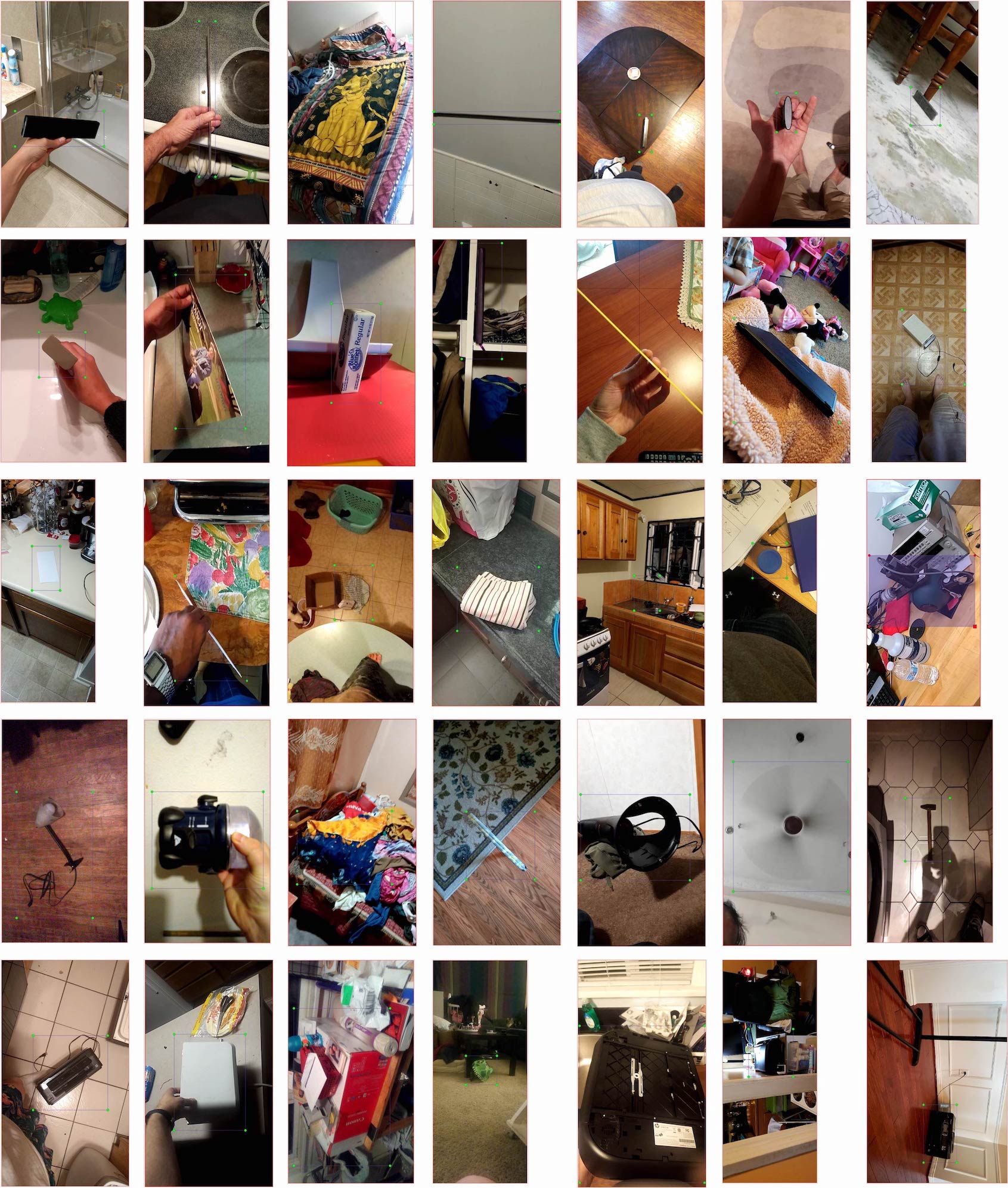}
    \caption{A selection of challenging objects that are hard to be recognized by humans (continued from above). Can you guess the category of the annotated objects in these images? Keys are as follows:  \newline
    row 1: \emph{(remote control, ruler, full sized towel, ruler, remote control, remote control, remote control)}, \newline
    row 2: \emph{(remote control, calendar, butter, bookend, ruler, tray, desk lamp)},   \newline
    row 3: \emph{(envelope, envelope, drying rack for dishes, full sized towel, drying rack for dishes, drinking cup, desk lamp)},  \newline
    row 4: \emph{(desk lamp, desk lamp, dress, tennis racket, fan, fan, hammer)},  \newline
    row 5: \emph{(printer, toaster, printer, helmet, printer, printer, printer)}
    }    
    \label{fig:Samples2}
\end{figure}

\section{Discussion and Conclusion}

Our investigation reveals that deep object recognition models perform significantly better when applied to isolated objects rather than scenes (around 20-30\% increase in performance). The reason behind this is two fold. First, there is less variability in single objects compared to scenes containing those objects. Second, deep models used here have been trained on ImageNet images which are less cluttered compared to the ObjectNet images. We anticipate that training models from scratch on large scale datasets that contain isolated objects will likely result in even higher accuracy.

Assuming around 30\% increase in performance (at best) over the Barbu {\em et al.}'s results using bounding boxes, still leaves a large gap of at least 15\% between ImageNet and ObjectNet performances which means that ObjectNet is significantly much harder. It covers a wider range of variations than ImageNet including object instances, viewpoints, rotations, occlusions, etc which pushes the limits of object recognition in both humans and machines. Hence, despite its limitations and biases, ObjectNet is indeed a great resource to test models in realistic situations.

Throughout the annotation process of ObjectNet images, we came across the following observations:
\begin{itemize}
    \item Some objects look very different when they are in motion (\eg the fan in Fig.~\ref{fig:Samples2}; row 4)

    \item Some objects appear different under the shadow of other objects (\eg the hammer in Fig.~\ref{fig:Samples2}; row 4)

    \item Some object instances look very different from the typical instances in the same class (\eg the helmet in Fig.~\ref{fig:Samples2}; row 5, the orange in Fig.~\ref{fig:Samples1}; row 5)
    
    \item Some objects can be recognized only by reading their labels (\eg the pet food container in Fig.~\ref{fig:Samples1}; row 2)

    \item Some images have wrong labels (\eg the pillow in Fig.~\ref{fig:Samples1}; row 2, the skirt in Fig.~\ref{fig:Samples1}; row 1, the tray in Fig.~\ref{fig:Samples2}; row 2.)
    
    \item Some objects are extremely difficult to be recognized by humans (\eg the tennis racket in Fig.~\ref{fig:Samples2}; row 4, the shovel in Fig.~\ref{fig:Samples1}; row 4, and the tray in Fig.~\ref{fig:Samples1}; row 1)
    
    \item In many images, objects are occluded by hands holding them (\eg the sock and the shovel in Fig.~\ref{fig:Samples1}; row 4)
    
    \item Some objects are hard to recognize in dim light (\eg the printer in Fig.~\ref{fig:Samples1}; row 2)
    
    \item Some categories are often confused with other categories, for example: \textit{(bath towel, bed sheet, full sized towel, dishrag or hand towel)}, \textit{(sandal, dress shoe (men), running shoe)}, \textit{(t-shirt, dress, sweater, suit jacket, skirt)}, \textit{(ruler, spatula, pen, match)}, \textit{(padlock, combination lock)}, and \textit{(envelope, letter)}.    
\end{itemize}

We foresee at least four directions for future work in this area: 
\begin{enumerate}
    \item First, it would be interesting to 
see how well state of the art object detectors (\eg Faster RCNN~\cite{ren2015faster}) perform on this dataset (\eg over classes overlapped with the MSCOCO dataset). We expect a big drop in detection performance since recognition is still the main bottleneck in object detection~\cite{borji2019empirical}.

\item Second, measuring human performance on ObjectNet dataset will provide a baseline for gauging model performance. Barbu \etal report a human performance of around 95\% on this dataset (via a pilot study) when subjects are asked to mention the objects that are present in the scene. This task, however, is different than recognizing isolated objects out of context similar to the regime that is considered here (\ie similar to rapid scene categorization tasks). In addition, error patterns of models and humans (rather than just raw accuracy measures) will inform us about the mechanisms of object recognition in both humans and machines. It could be that models work in a completely different fashion than the human visual system. For instance, unlike CNNs, we are able to discern the foreground from the image background during recognition. This hints towards an interplay and feedback loop between recognition and segmentation that is currently missing in CNNs. Finally, we are invariant to image transformation only partially, therefore it may not make sense to desire models that are fully invariant (\eg invariance to 360 degree in-plane rotation).

\item Third, and related to the second, is the role of context in object recognition. Context is a two-edge sword. On the one hand, using it may lead to relying on trivial correlations that may not always happen at the test time. For example, relying on the fact that a keyboard always appears next to a monitor, may lead to occasional failures when a model is tested on an isolated keyboard. A better example is adversarial patches~\cite{brown2017adversarial} where training a model on objects augmented with small random patches forces the model to rely on those features and hence get fooled when tested on other objects with the same patch. On the other hand, completely discarding the context also is a not wise for two reasons. First, the great success of CNNs for object recognition and scene segmentation is attributed to their ability to exploit visual context. Second, as humans we also heavily rely on surrounding visual context. Future research, should further investigate the role of context in object recognition and how best to exploit it. 

\item Fourth, to operationalize the third item above, we propose a new task which is recognizing objects in cluttered scenes containing multiple objects (See~\cite{borji2019empirical} for an example study). This task resembles object detection but it is actually different. First, here the ground truth object bounding boxes are given and the task is to correctly label them (\ie no need to find objects). Second, the evaluation measure is accuracy which is easier to interpret than mean average precision. This task is also different than the current object recognition setup in which context is usually discarded. Existing object detection datasets, such as MS COCO~\cite{lin2014microsoft}, can be utilized to evaluate models built for solving this task (\ie instead of detecting objects the aim is to recognize isolated objects confined in bounding boxes).
\end{enumerate}

{\small
\bibliographystyle{IEEEtran}
\bibliography{refs}

\begin{thebibliography}{10}
\providecommand{\url}[1]{#1}
\csname url@samestyle\endcsname
\providecommand{\newblock}{\relax}
\providecommand{\bibinfo}[2]{#2}
\providecommand{\BIBentrySTDinterwordspacing}{\spaceskip=0pt\relax}
\providecommand{\BIBentryALTinterwordstretchfactor}{4}
\providecommand{\BIBentryALTinterwordspacing}{\spaceskip=\fontdimen2\font plus
\BIBentryALTinterwordstretchfactor\fontdimen3\font minus
  \fontdimen4\font\relax}
\providecommand{\BIBforeignlanguage}[2]{{%
\expandafter\ifx\csname l@#1\endcsname\relax
\typeout{** WARNING: IEEEtran.bst: No hyphenation pattern has been}%
\typeout{** loaded for the language `#1'. Using the pattern for}%
\typeout{** the default language instead.}%
\else
\language=\csname l@#1\endcsname
\fi
#2}}
\providecommand{\BIBdecl}{\relax}
\BIBdecl

\bibitem{lecun1998gradient}
Y.~LeCun, L.~Bottou, Y.~Bengio, and P.~Haffner, ``Gradient-based learning
  applied to document recognition,'' \emph{Proceedings of the IEEE}, vol.~86,
  no.~11, pp. 2278--2324, 1998.

\bibitem{krizhevsky2012imagenet}
A.~Krizhevsky, I.~Sutskever, and G.~E. Hinton, ``Imagenet classification with
  deep convolutional neural networks,'' in \emph{Advances in neural information
  processing systems}, 2012, pp. 1097--1105.

\bibitem{azulay2019deep}
A.~Azulay and Y.~Weiss, ``Why do deep convolutional networks generalize so
  poorly to small image transformations?'' \emph{Journal of Machine Learning
  Research}, vol.~20, no. 184, pp. 1--25, 2019.

\bibitem{recht2019imagenet}
B.~Recht, R.~Roelofs, L.~Schmidt, and V.~Shankar, ``Do imagenet classifiers
  generalize to imagenet?'' \emph{arXiv preprint arXiv:1902.10811}, 2019.

\bibitem{goodfellow2014explaining}
I.~J. Goodfellow, J.~Shlens, and C.~Szegedy, ``Explaining and harnessing
  adversarial examples,'' \emph{arXiv preprint arXiv:1412.6572}, 2014.

\bibitem{imagenet}
O.~Russakovsky, J.~Deng, H.~Su, J.~Krause, S.~Satheesh, S.~Ma, Z.~Huang,
  A.~Karpathy, A.~Khosla, M.~Bernstein, A.~C. Berg, and L.~Fei-Fei, ``{ImageNet
  Large Scale Visual Recognition Challenge},'' \emph{International Journal of
  Computer Vision (IJCV)}, vol. 115, no.~3, pp. 211--252, 2015.

\bibitem{krizhevsky2009learning}
A.~Krizhevsky, G.~Hinton \emph{et~al.}, ``Learning multiple layers of features
  from tiny images,'' 2009.

\bibitem{lecun2004learning}
Y.~LeCun, F.~J. Huang, and L.~Bottou, ``Learning methods for generic object
  recognition with invariance to pose and lighting,'' in \emph{Proceedings of
  the 2004 IEEE Computer Society Conference on Computer Vision and Pattern
  Recognition, 2004. CVPR 2004.}, vol.~2.\hskip 1em plus 0.5em minus
  0.4em\relax IEEE, 2004, pp. II--104.

\bibitem{borji2016ilab}
A.~Borji, S.~Izadi, and L.~Itti, ``ilab-20m: A large-scale controlled object
  dataset to investigate deep learning,'' in \emph{Proceedings of the IEEE
  Conference on Computer Vision and Pattern Recognition}, 2016, pp. 2221--2230.

\bibitem{barbu2019objectnet}
A.~Barbu, D.~Mayo, J.~Alverio, W.~Luo, C.~Wang, D.~Gutfreund, J.~Tenenbaum, and
  B.~Katz, ``Objectnet: A large-scale bias-controlled dataset for pushing the
  limits of object recognition models,'' in \emph{Advances in Neural
  Information Processing Systems}, 2019, pp. 9448--9458.

\bibitem{lin2014microsoft}
T.-Y. Lin, M.~Maire, S.~Belongie, J.~Hays, P.~Perona, D.~Ramanan,
  P.~Doll{\'a}r, and C.~L. Zitnick, ``Microsoft coco: Common objects in
  context,'' in \emph{European conference on computer vision}.\hskip 1em plus
  0.5em minus 0.4em\relax Springer, 2014, pp. 740--755.

\bibitem{singh2018analysis}
B.~Singh and L.~S. Davis, ``An analysis of scale invariance in object detection
  snip,'' in \emph{Proceedings of the IEEE conference on computer vision and
  pattern recognition}, 2018, pp. 3578--3587.

\bibitem{simonyan2014very}
K.~Simonyan and A.~Zisserman, ``Very deep convolutional networks for
  large-scale image recognition,'' \emph{arXiv preprint arXiv:1409.1556}, 2014.

\bibitem{szegedy2015going}
C.~Szegedy, W.~Liu, Y.~Jia, P.~Sermanet, S.~Reed, D.~Anguelov, D.~Erhan,
  V.~Vanhoucke, and A.~Rabinovich, ``Going deeper with convolutions,'' in
  \emph{Proceedings of the IEEE conference on computer vision and pattern
  recognition}, 2015, pp. 1--9.

\bibitem{resnet}
K.~He, X.~Zhang, S.~Ren, and J.~Sun, ``Deep residual learning for image
  recognition,'' in \emph{Proceedings of the IEEE Conference on Computer Vision
  and Pattern Recognition}, 2016, pp. 770--778.

\bibitem{szegedy2016rethinking}
C.~Szegedy, V.~Vanhoucke, S.~Ioffe, J.~Shlens, and Z.~Wojna, ``Rethinking the
  inception architecture for computer vision,'' in \emph{Proceedings of the
  IEEE conference on computer vision and pattern recognition}, 2016, pp.
  2818--2826.

\bibitem{tan2019mnasnet}
M.~Tan, B.~Chen, R.~Pang, V.~Vasudevan, M.~Sandler, A.~Howard, and Q.~V. Le,
  ``Mnasnet: Platform-aware neural architecture search for mobile,'' in
  \emph{Proceedings of the IEEE Conference on Computer Vision and Pattern
  Recognition}, 2019, pp. 2820--2828.

\bibitem{ren2015faster}
S.~Ren, K.~He, R.~Girshick, and J.~Sun, ``Faster r-cnn: Towards real-time
  object detection with region proposal networks,'' in \emph{Advances in neural
  information processing systems}, 2015, pp. 91--99.

\bibitem{borji2019empirical}
A.~Borji and S.~M. Iranmanesh, ``Empirical upper-bound in object detection and
  more,'' \emph{arXiv preprint arXiv:1911.12451}, 2019.

\bibitem{brown2017adversarial}
T.~B. Brown, D.~Mané, A.~Roy, M.~Abadi, and J.~Gilmer, ``Adversarial patch,''
  2017.

\end{thebibliography}
}


\section{Appendix}
Here, we show the easiest and hardest objects for the ResNet-152 model over some categories.

\begin{figure}[!htbp]
\centering

\subfigure[Correctly classified; highest confidences]{\includegraphics[width=1\linewidth]{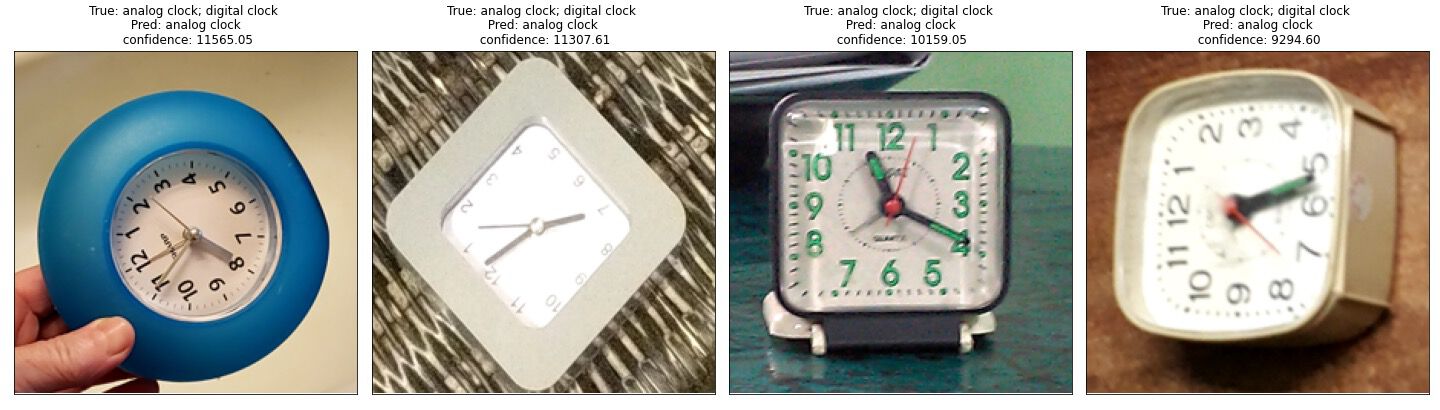}}
\vspace{-10pt}
\subfigure[Correctly classified; lowest confidences]{\includegraphics[width=1\linewidth]{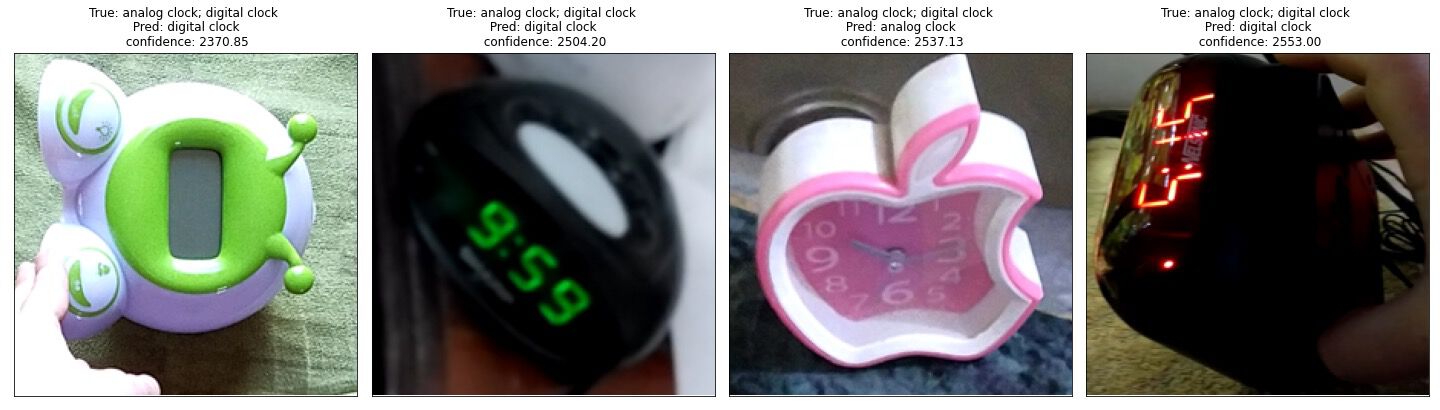}}\vspace{50pt}
\subfigure[Misclassified; highest confidences]{\includegraphics[width=1\linewidth]{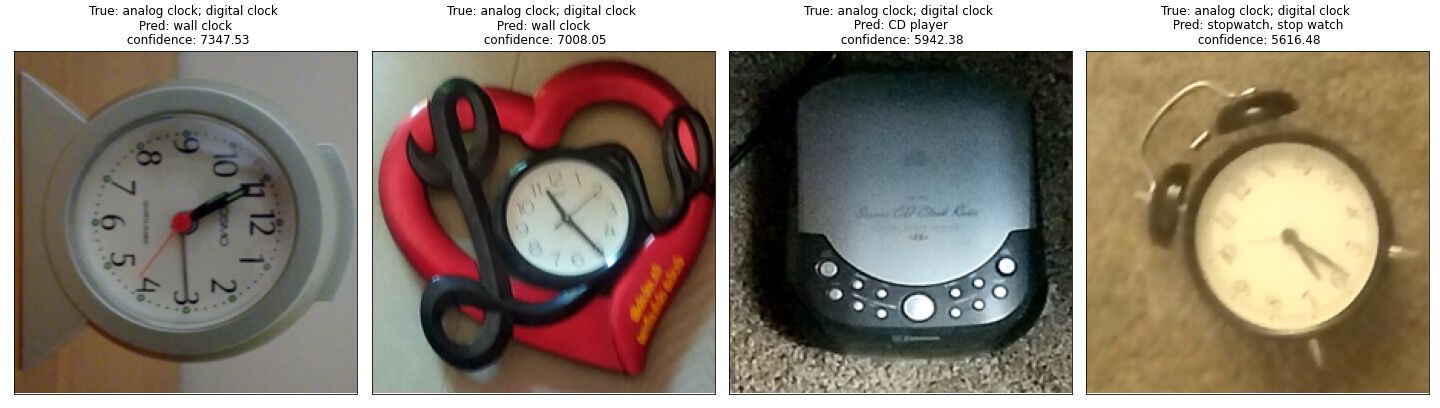}}\vspace{-10pt}
\subfigure[Misclassified; lowest confidences]{\includegraphics[width=1\linewidth]{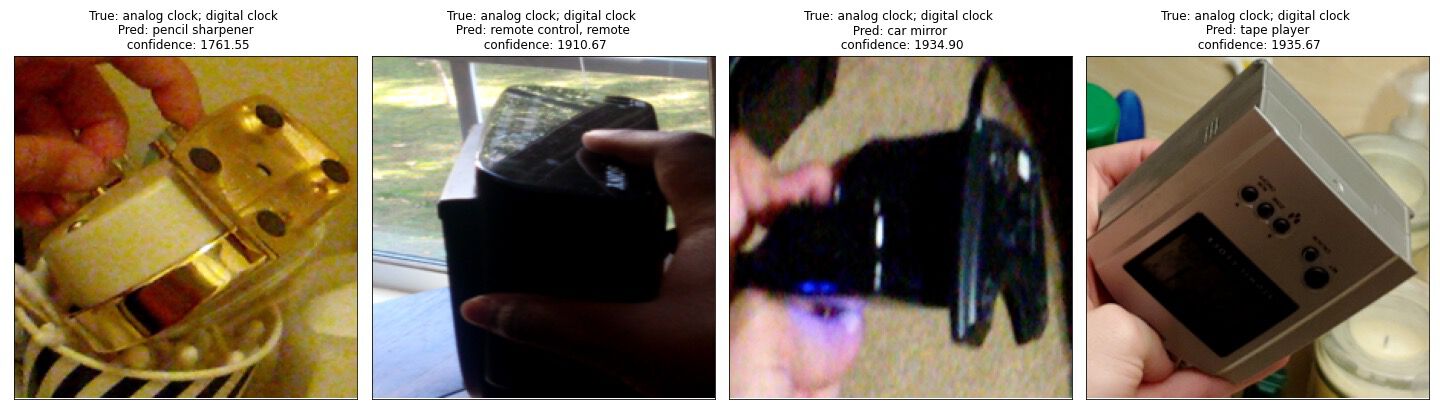}}

\caption{Correctly classified and misclassified examples from the Alarm clock class by the ResNet model.}
\label{fig:alarmclock}
\end{figure}

\begin{figure}
\centering

\subfigure[Correctly classified; highest confidences]{\includegraphics[width=1\linewidth]{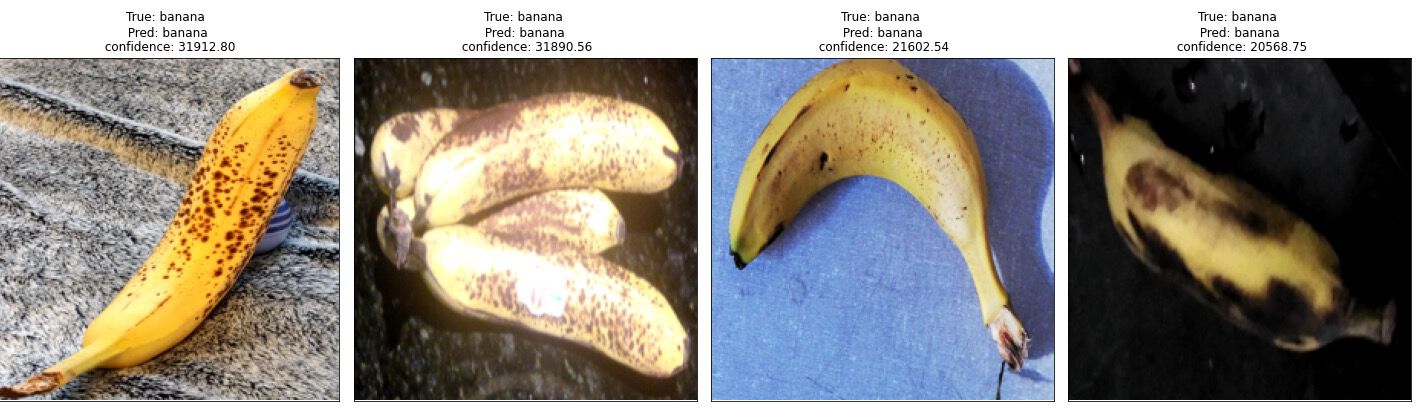}}
\vspace{-10pt}
\subfigure[Correctly classified; lowest confidences]{\includegraphics[width=1\linewidth]{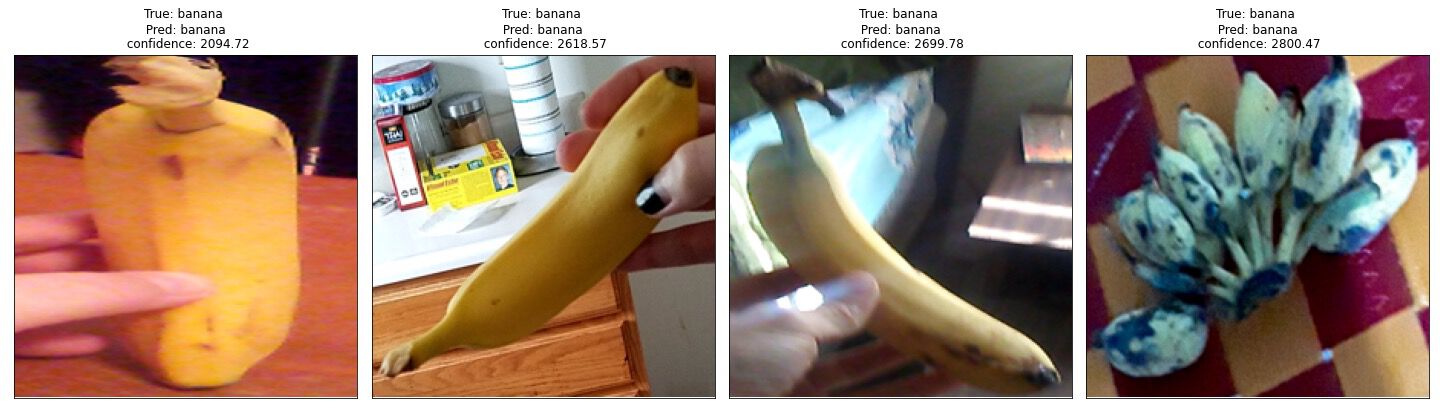}}\vspace{50pt}
\subfigure[Misclassified; highest confidences]{\includegraphics[width=1\linewidth]{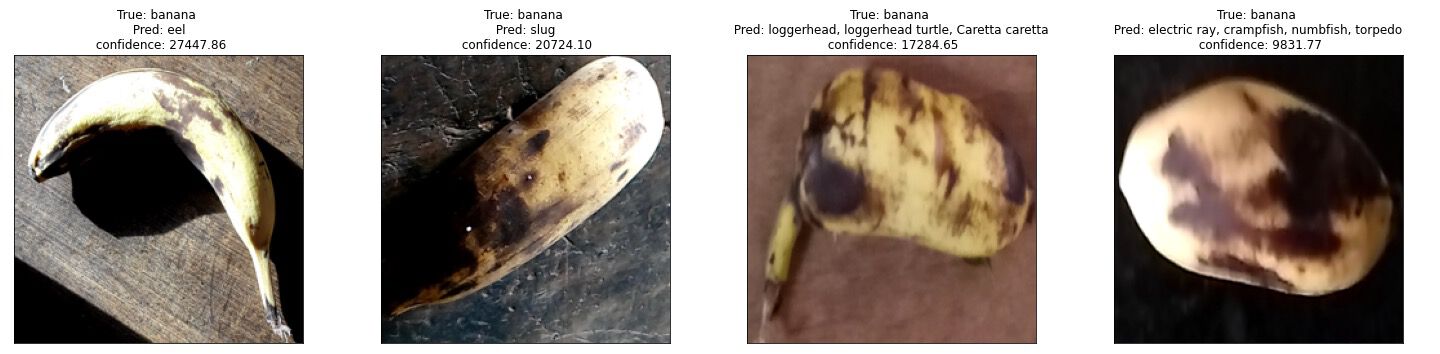}}\vspace{-10pt}
\subfigure[Misclassified; lowest confidences]{\includegraphics[width=1\linewidth]{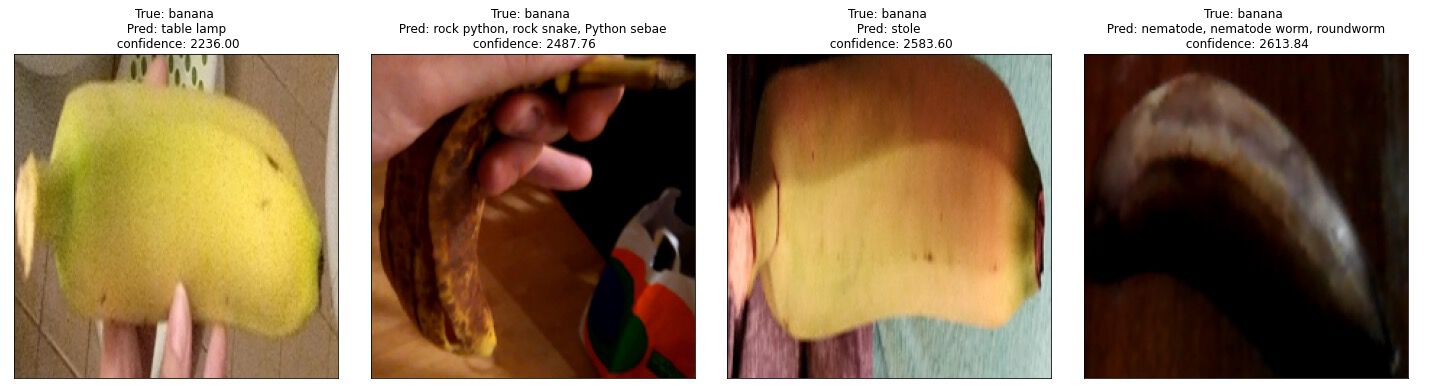}}

\caption{Correctly classified and misclassified examples from the Banana class by the ResNet model.}
\label{fig:alarmclock}
\end{figure}

\begin{figure}
\centering
\subfigure[Correctly classified; highest confidences]{\includegraphics[width=1\linewidth]{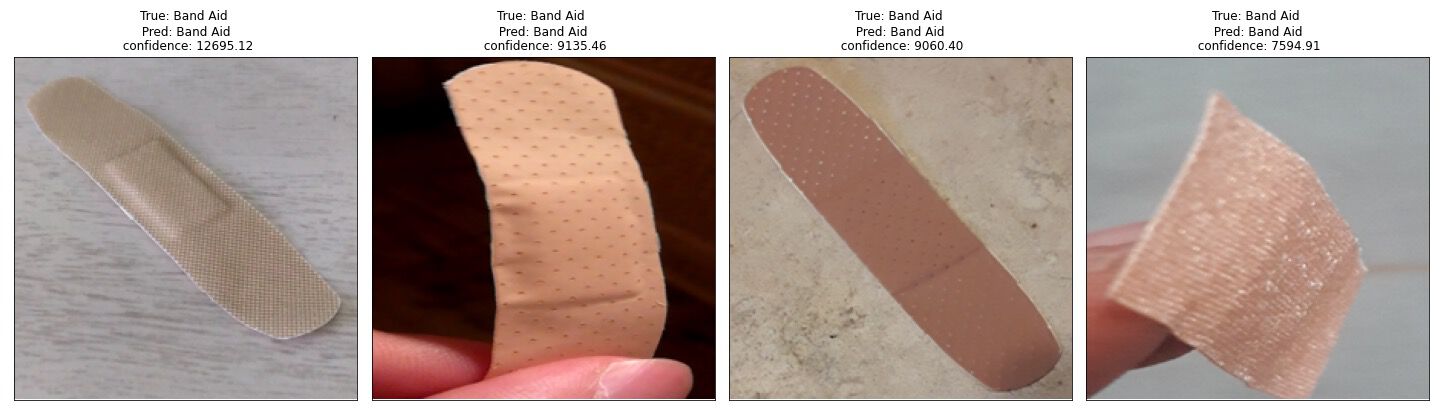}}
\vspace{-10pt}
\subfigure[Correctly classified; lowest confidences]{\includegraphics[width=1\linewidth]{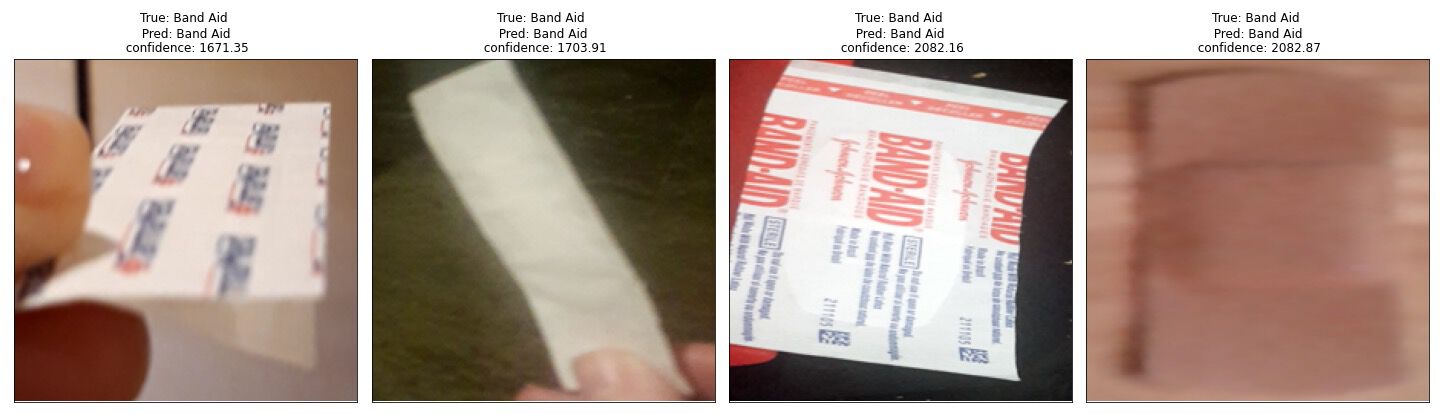}}\vspace{50pt}
\subfigure[Misclassified; highest confidences]{\includegraphics[width=1\linewidth]{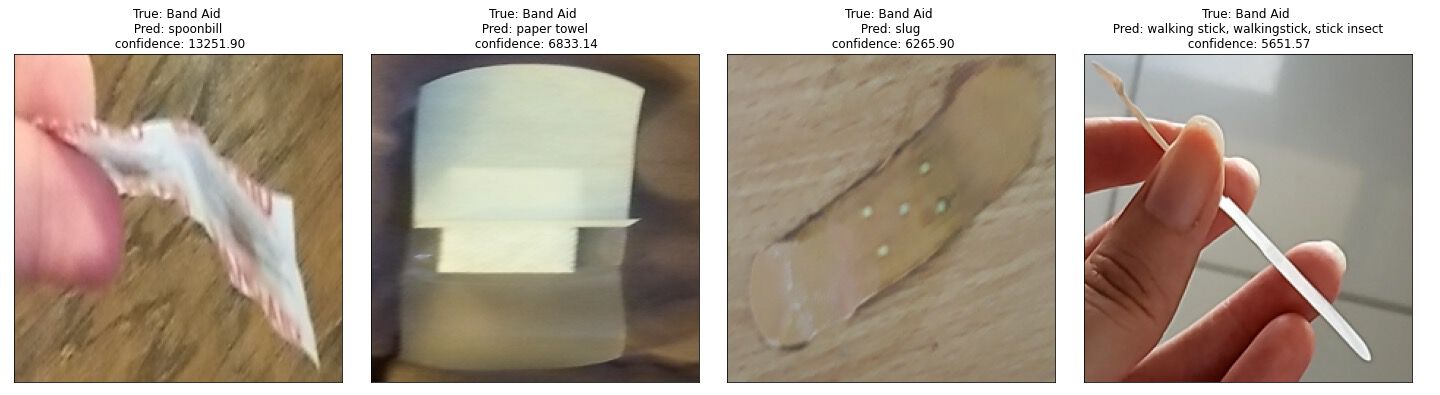}}\vspace{-10pt}
\subfigure[Misclassified; lowest confidences]{\includegraphics[width=1\linewidth]{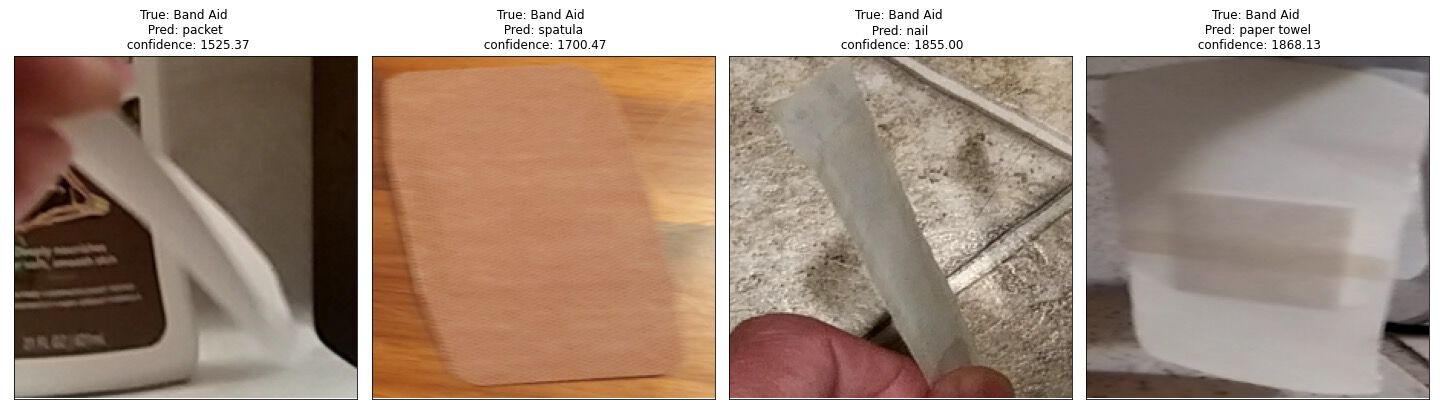}}
\caption{Correctly classified and misclassified examples from the Band-Aid class by the ResNet model.}
\label{fig:BandAid}
\end{figure}

\begin{figure}
\centering
\subfigure[Correctly classified; highest confidences. Only three benchs were correctly classified. ]{\includegraphics[width=1\linewidth]{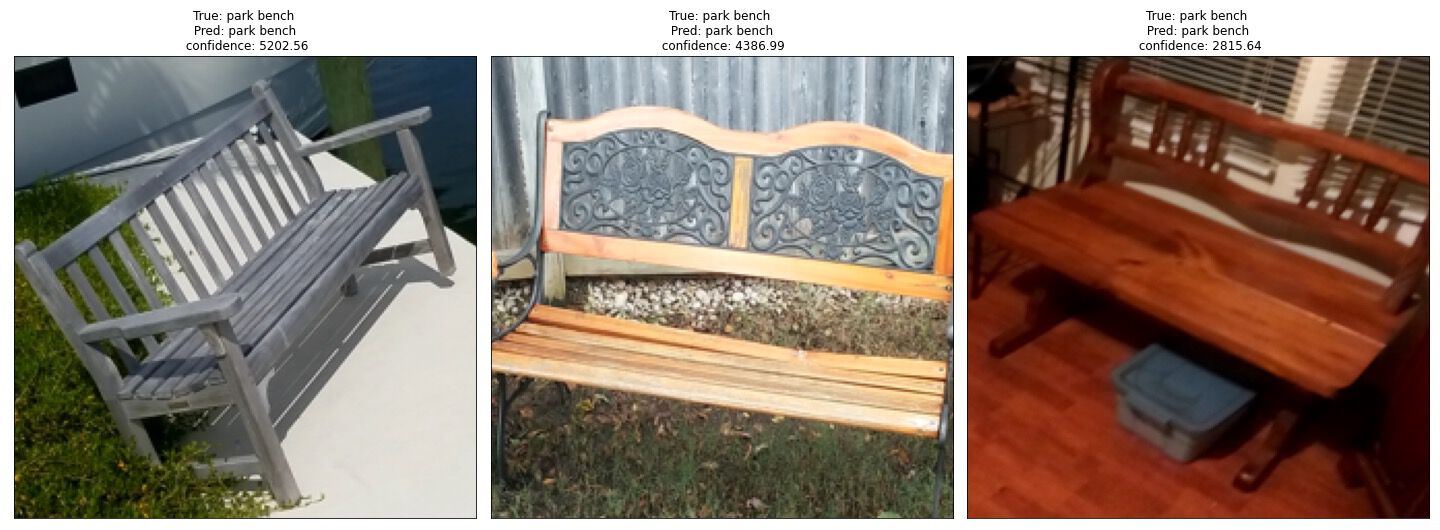}}
\vspace{50pt} \\
\subfigure[Misclassified; highest confidences]{\includegraphics[width=1\linewidth]{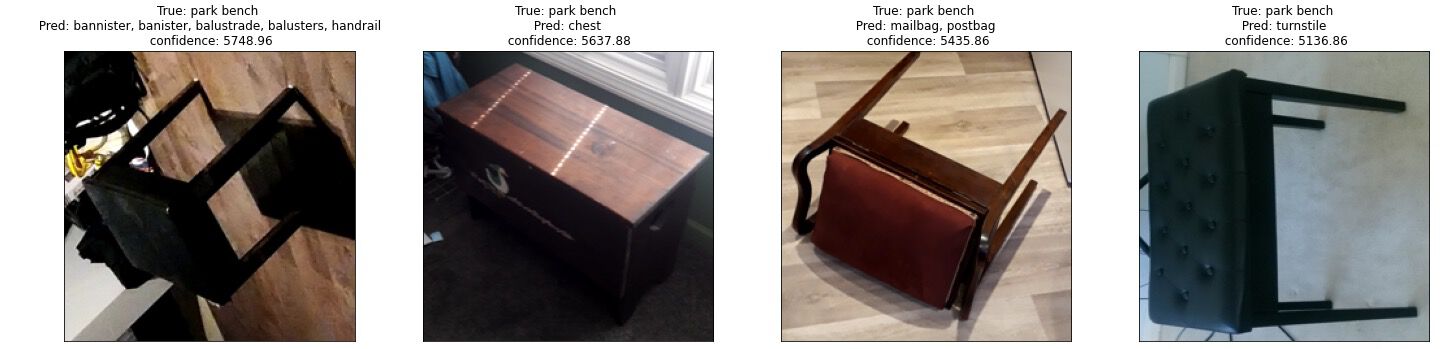}}\vspace{-10pt}
\subfigure[Misclassified; lowest confidences]{\includegraphics[width=1\linewidth]{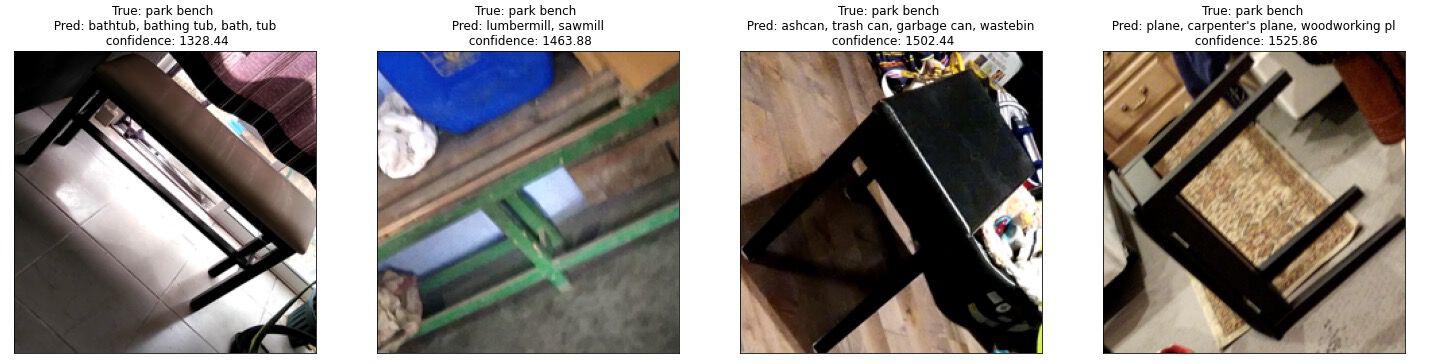}}
\caption{Correctly classified and misclassified examples from the Bench class by the ResNet model.}
\label{fig:Bench}
\end{figure}

\clearpage

\begin{figure}
\centering
\subfigure[Correctly classified; highest confidences. Only two plates were correctly classified.]{\includegraphics[width=1\linewidth]{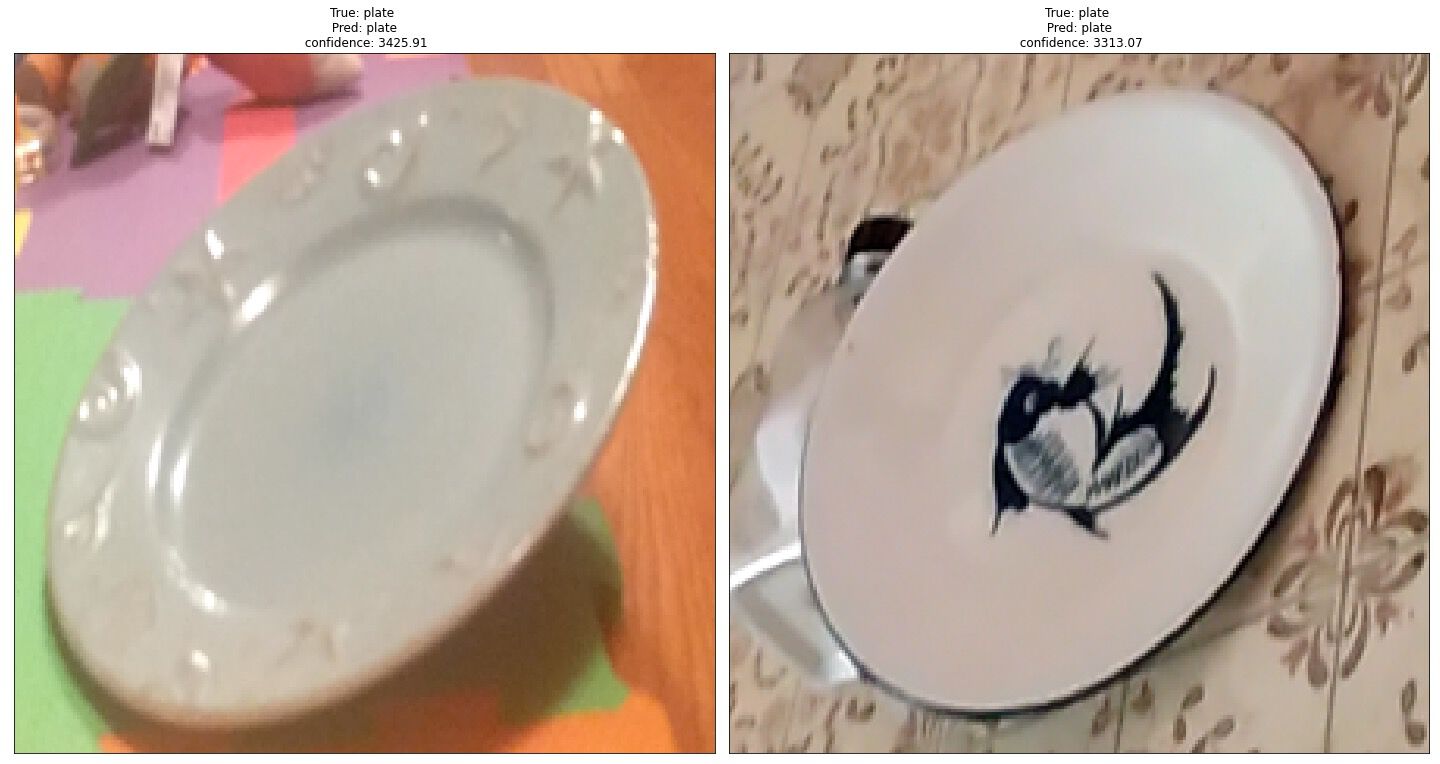}} \vspace{50pt} \\
\subfigure[Misclassified; highest confidences]{\includegraphics[width=1\linewidth]{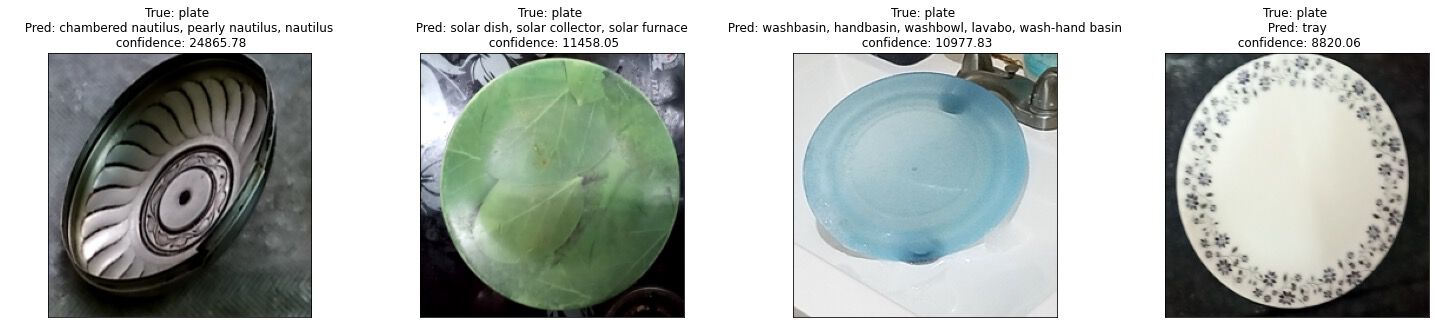}}\vspace{-10pt}
\subfigure[Misclassified; lowest confidences]{\includegraphics[width=1\linewidth]{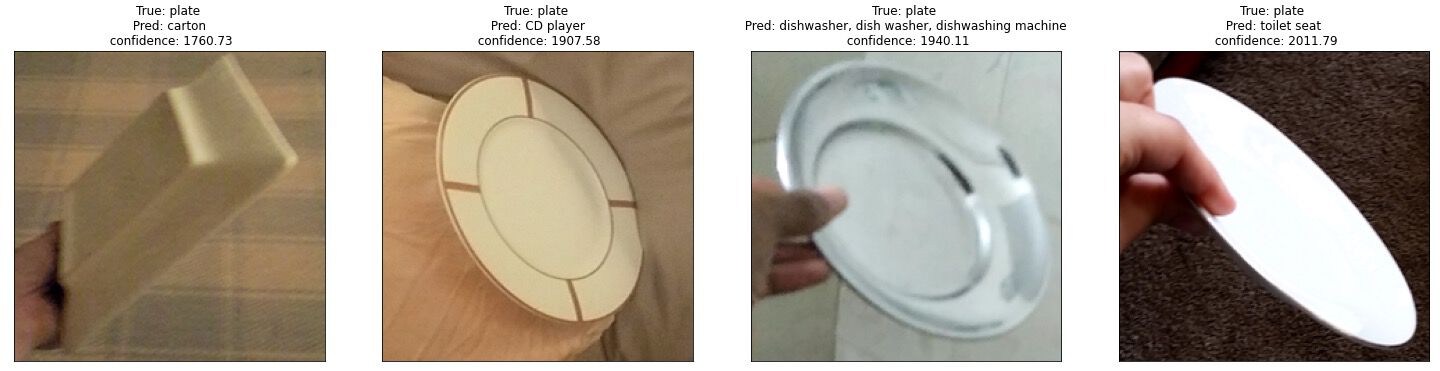}}\caption{Correctly classified and misclassified examples from the Plate class by the ResNet model.}
\label{fig:Plate}
\end{figure}

\clearpage

\begin{figure}
\centering
\subfigure[Correctly classified; highest confidences]{\includegraphics[width=1\linewidth]{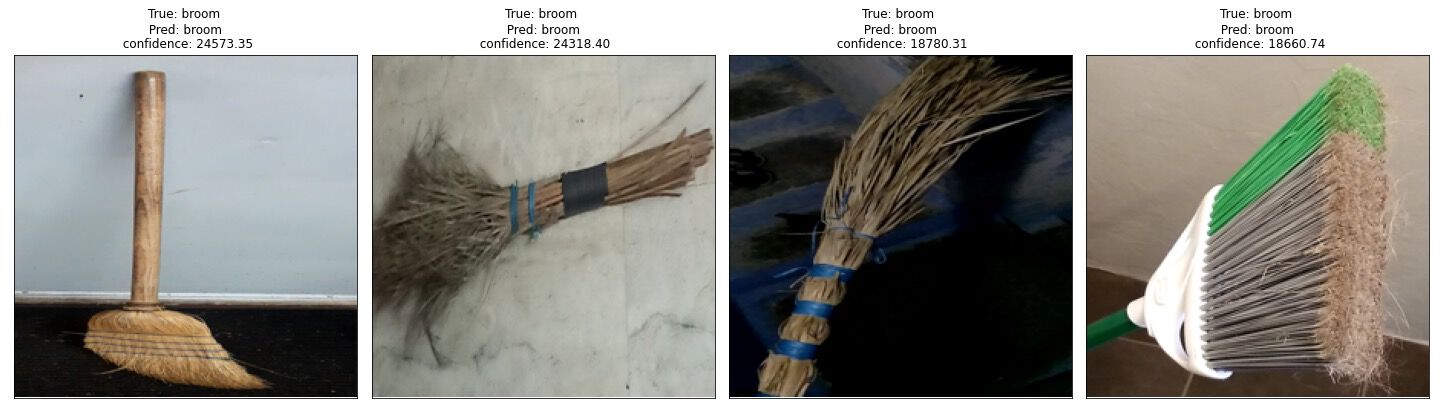}}
\vspace{-10pt}
\subfigure[Correctly classified; lowest confidences]{\includegraphics[width=1\linewidth]{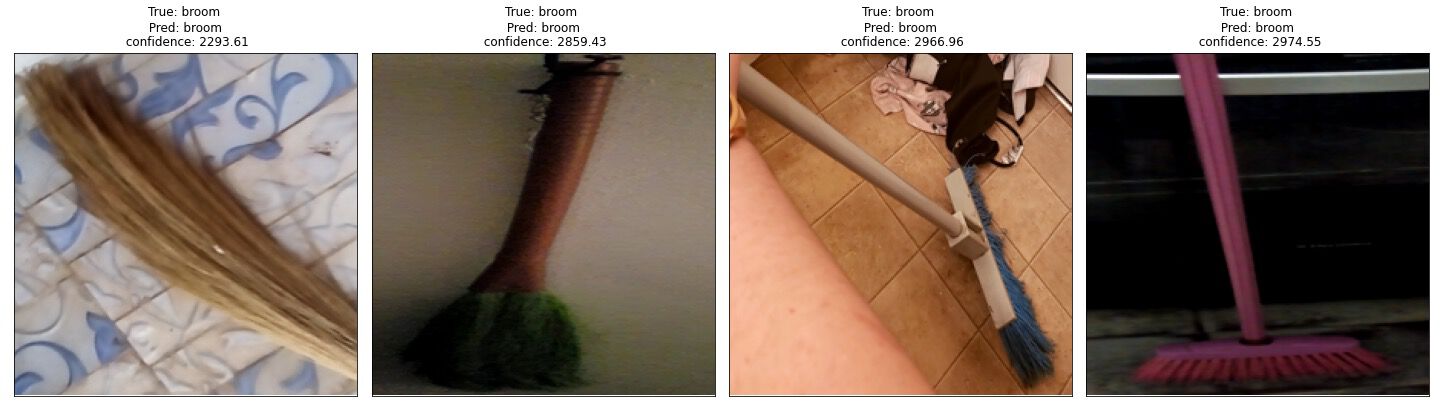}}\vspace{50pt}
\subfigure[Misclassified; highest confidences]{\includegraphics[width=1\linewidth]{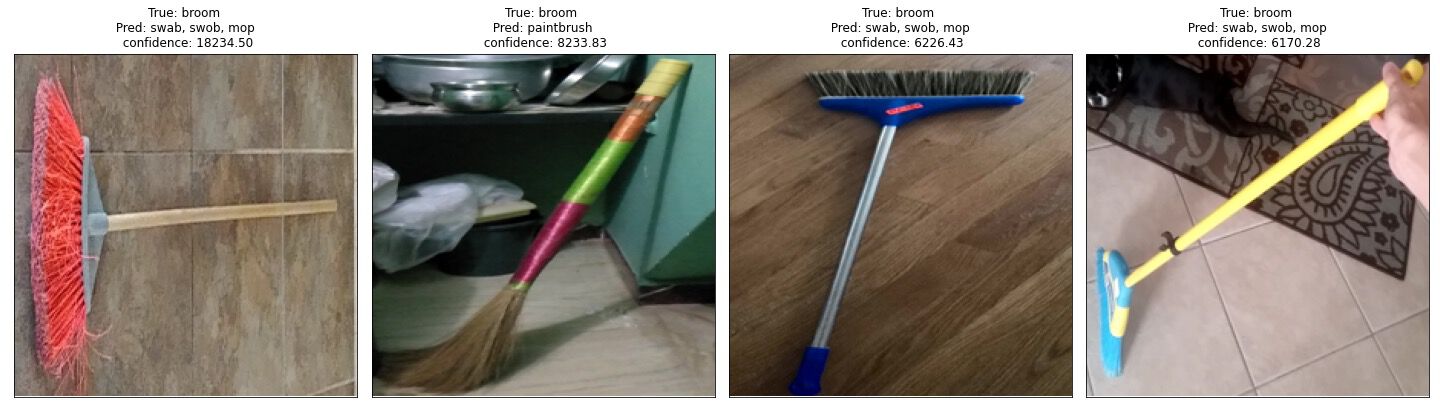}}\vspace{-10pt}
\subfigure[Misclassified; lowest confidences]{\includegraphics[width=1\linewidth]{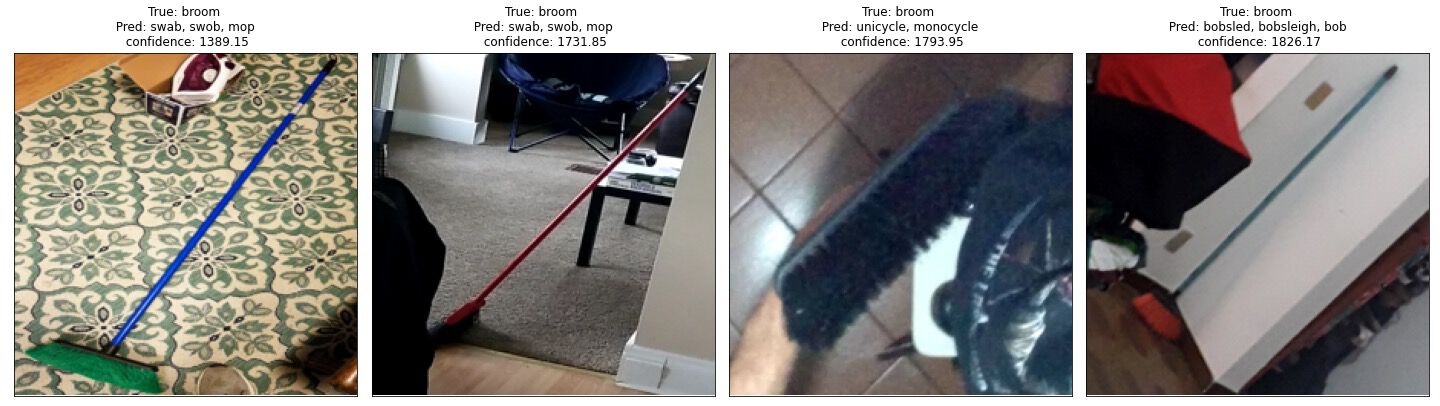}}
\caption{Correctly classified and misclassified examples from the Broom class by the ResNet model.}
\label{fig:Broom}
\end{figure}

\clearpage

\begin{figure}
\centering
\subfigure[Correctly classified; highest confidences]{\includegraphics[width=1\linewidth]{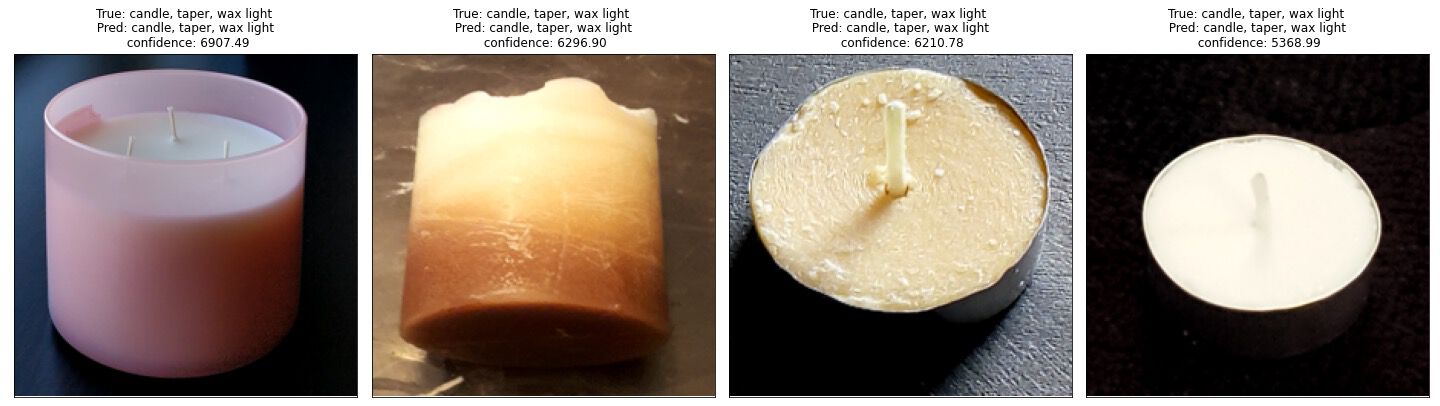}}
\vspace{-10pt}
\subfigure[Correctly classified; lowest confidences]{\includegraphics[width=1\linewidth]{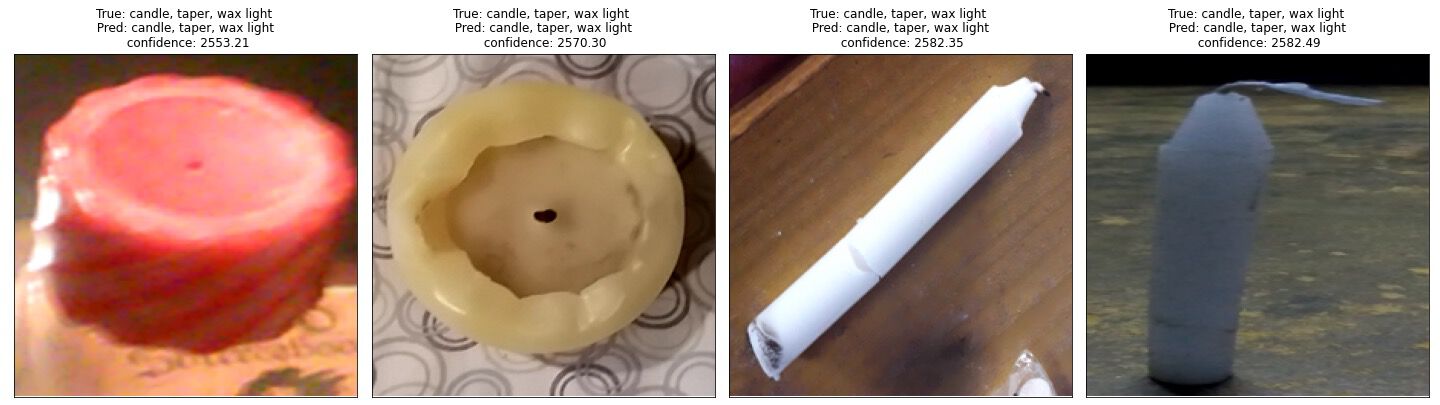}}\vspace{50pt}
\subfigure[Misclassified; highest confidences]{\includegraphics[width=1\linewidth]{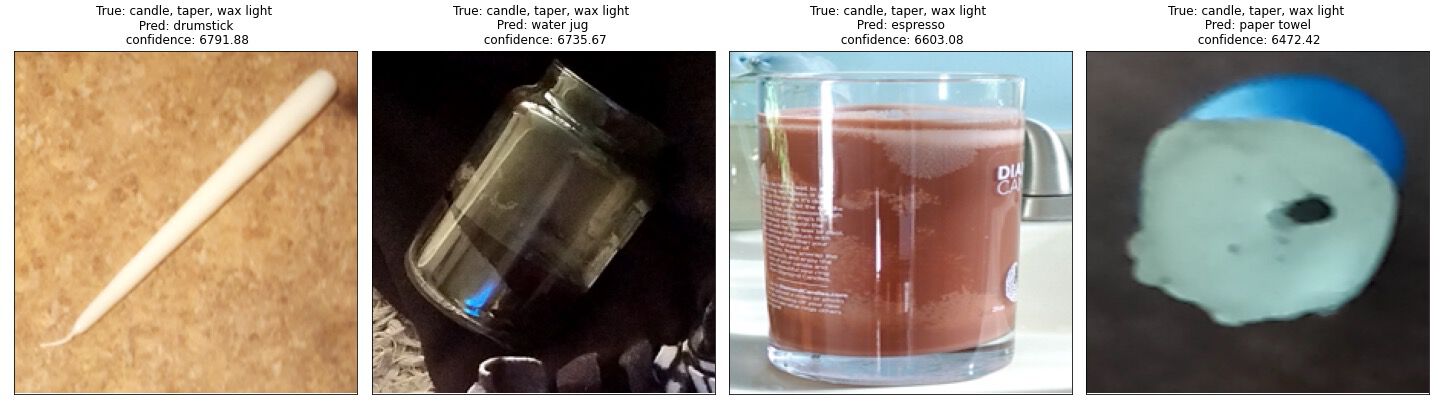}}\vspace{-10pt}
\subfigure[Misclassified; lowest confidences]{\includegraphics[width=1\linewidth]{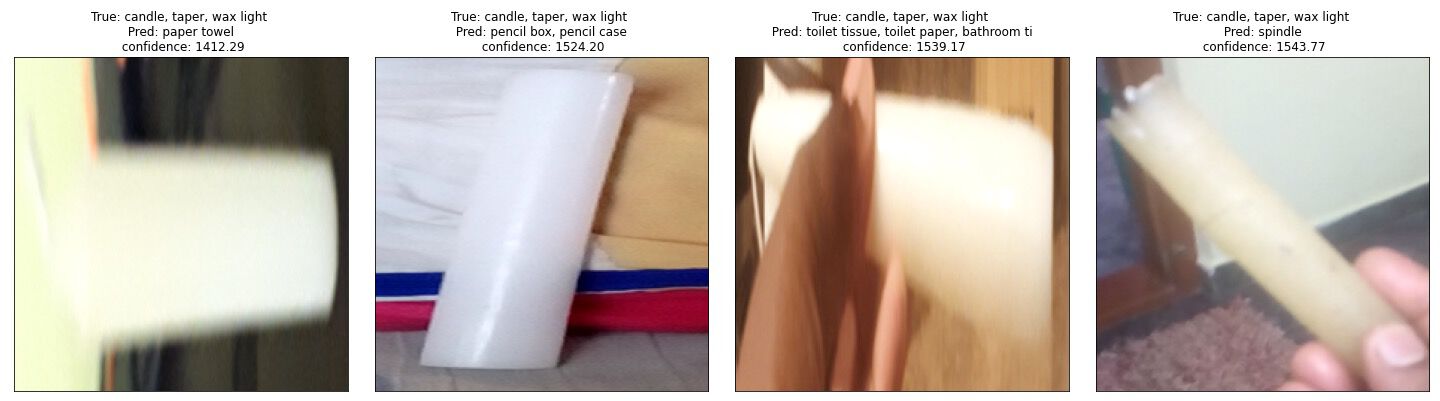}}
\caption{Correctly classified and misclassified examples from the Candle class by the ResNet model.}
\label{fig:Candle}
\end{figure}

\clearpage

\begin{figure}
\centering
\subfigure[Correctly classified; highest confidences]{\includegraphics[width=1\linewidth]{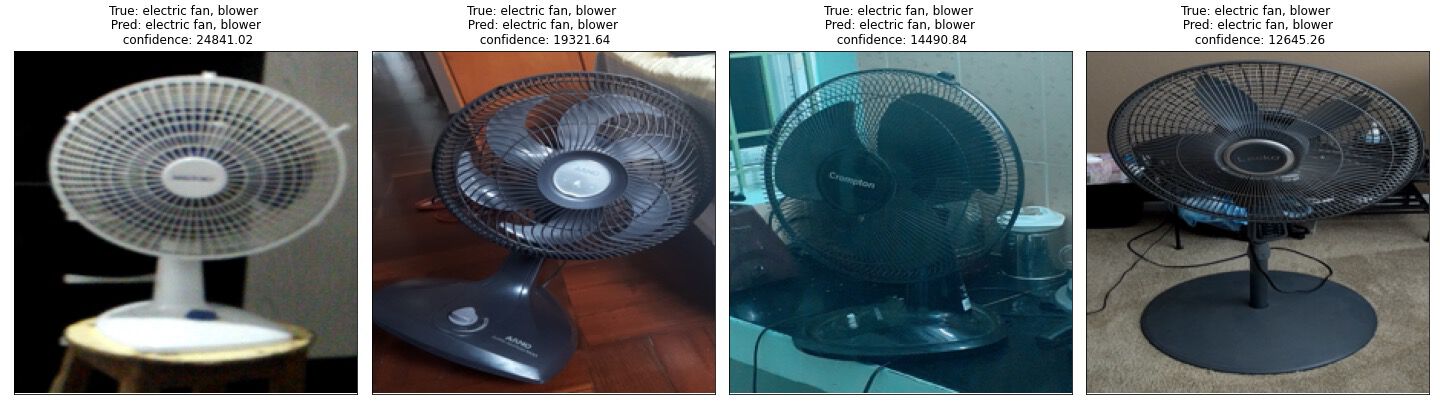}}
\vspace{-10pt}
\subfigure[Correctly classified; lowest confidences]{\includegraphics[width=1\linewidth]{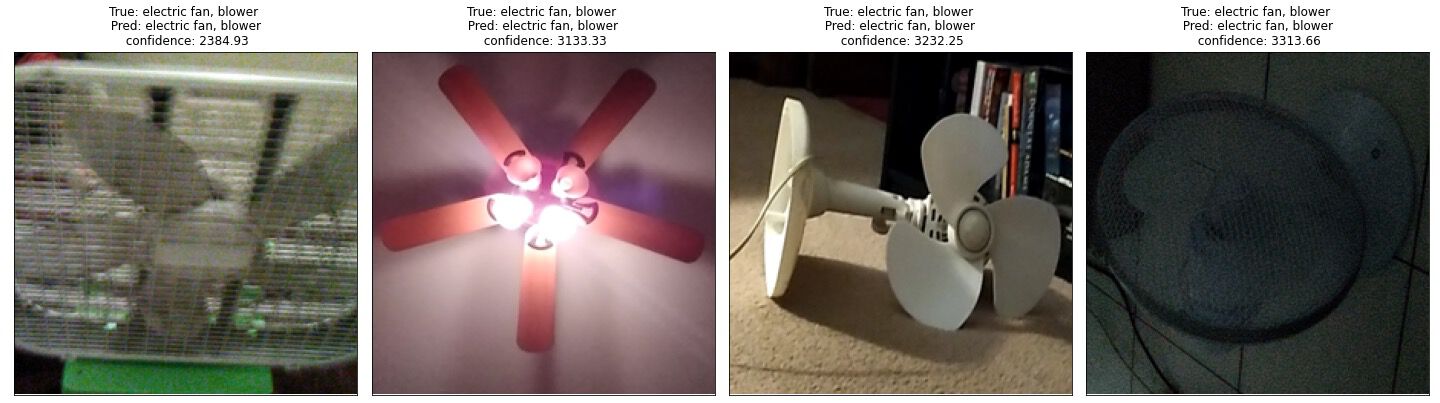}}\vspace{50pt}
\subfigure[Misclassified; highest confidences]{\includegraphics[width=1\linewidth]{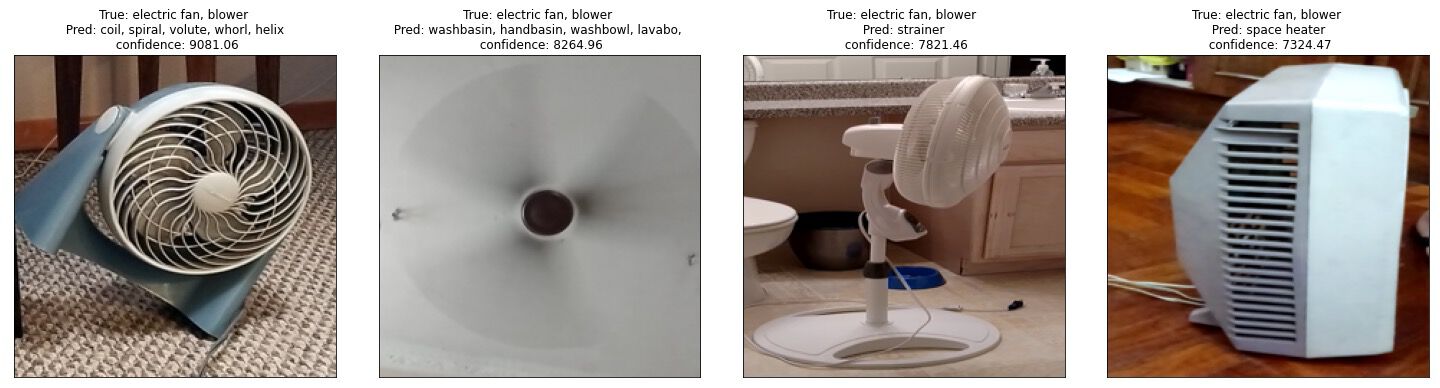}}\vspace{-10pt}
\subfigure[Misclassified; lowest confidences]{\includegraphics[width=1\linewidth]{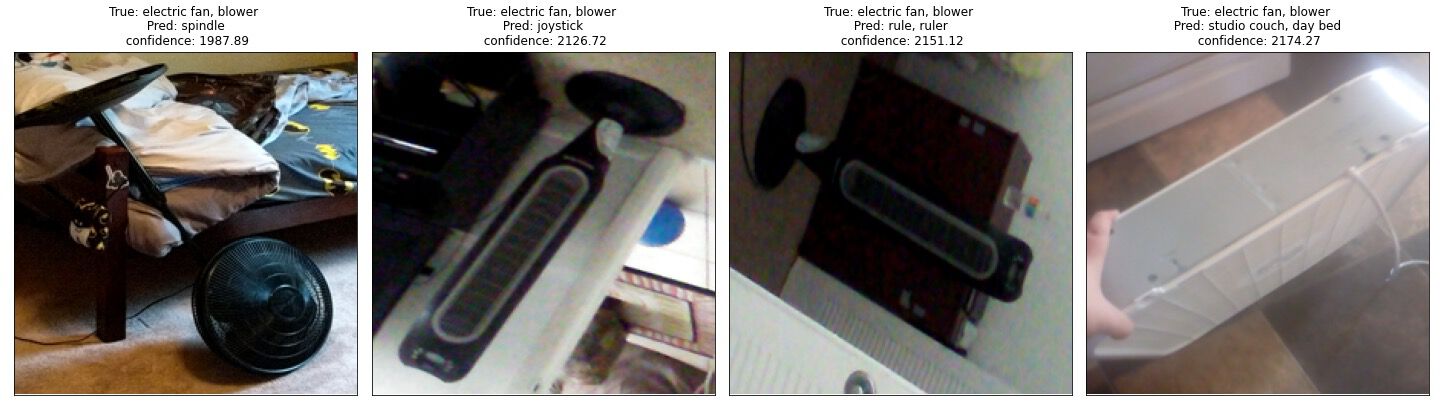}}
\caption{Correctly classified and misclassified examples from the Fan class by the ResNet model.}
\label{fig:Fan}
\end{figure}

\begin{figure}
\centering
\subfigure[Correctly classified; highest confidences]{\includegraphics[width=1\linewidth]{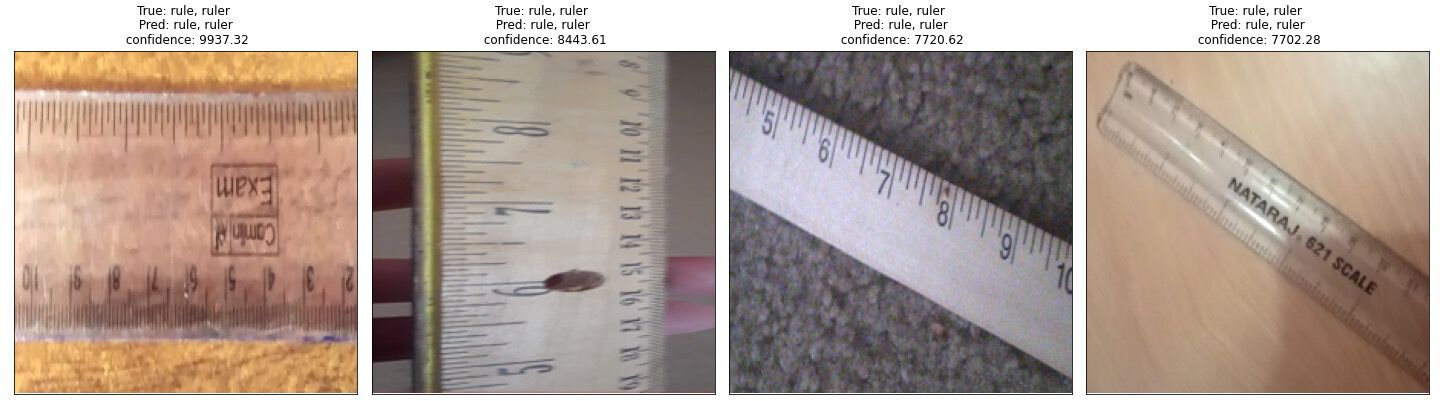}}
\vspace{-10pt}
\subfigure[Correctly classified; lowest confidences]{\includegraphics[width=1\linewidth]{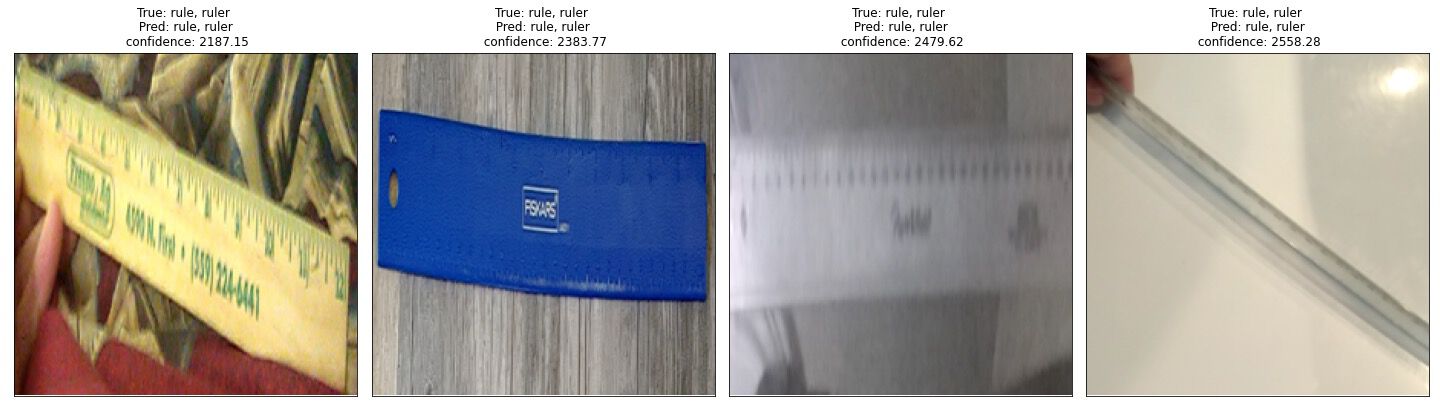}}\vspace{50pt}
\subfigure[Misclassified; highest confidences]{\includegraphics[width=1\linewidth]{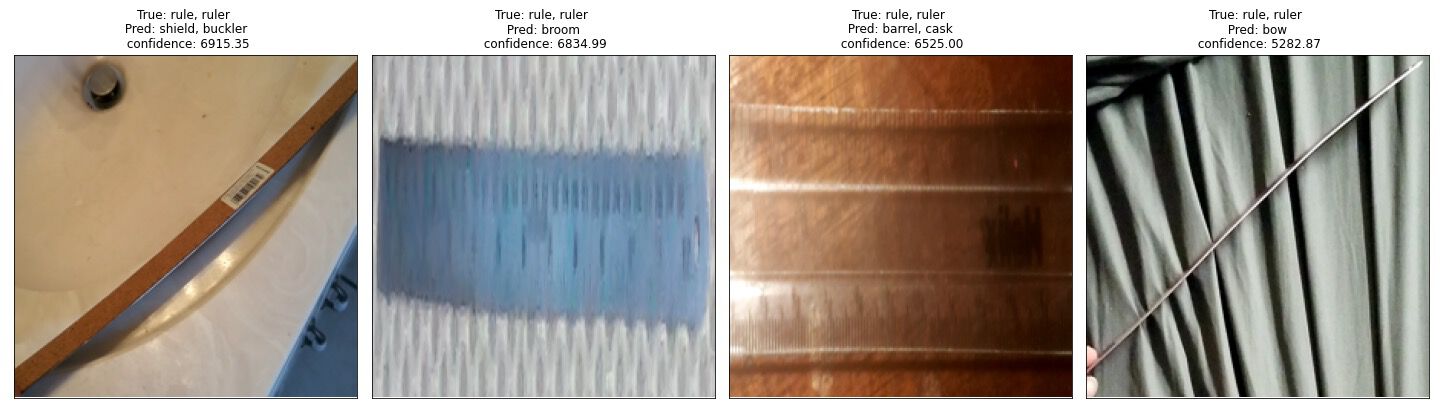}}\vspace{-10pt}
\subfigure[Misclassified; lowest confidences]{\includegraphics[width=1\linewidth]{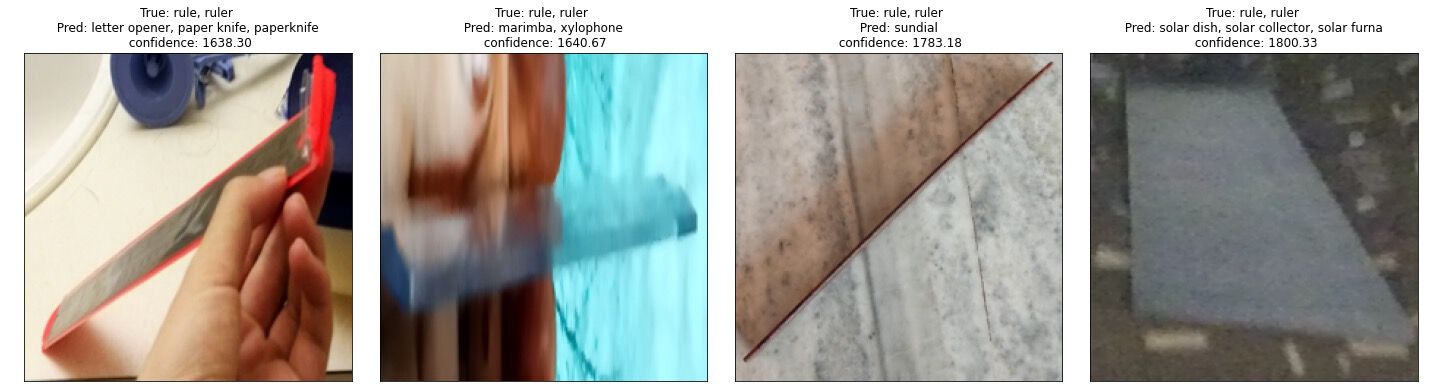}}
\caption{Correctly classified and misclassified examples from the Ruler class by the ResNet model.}
\label{fig:Ruler}
\end{figure}

\begin{figure}
\centering
\subfigure[Correctly classified; highest confidences]{\includegraphics[width=1\linewidth]{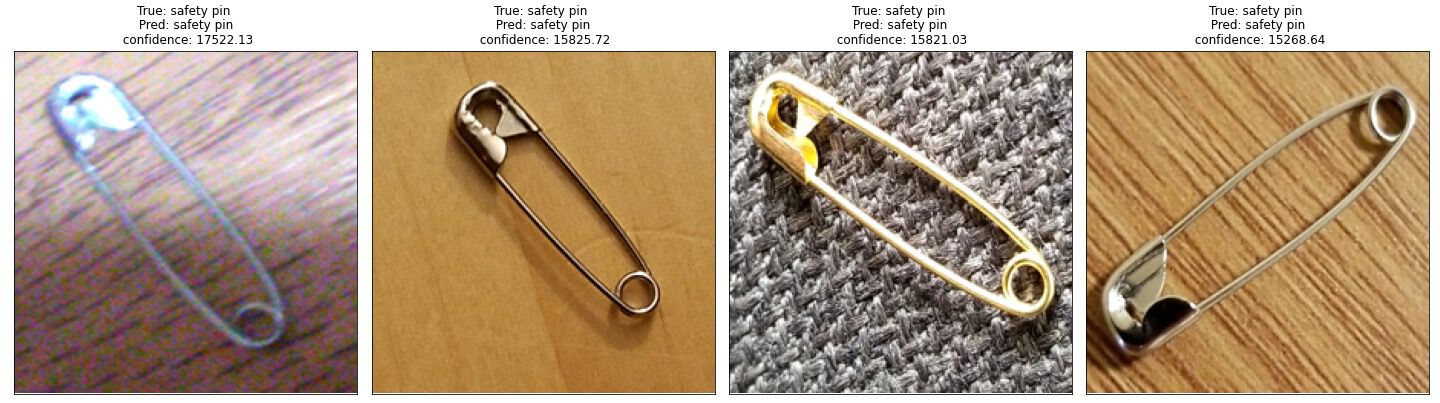}}
\vspace{-10pt}
\subfigure[Correctly classified; lowest confidences]{\includegraphics[width=1\linewidth]{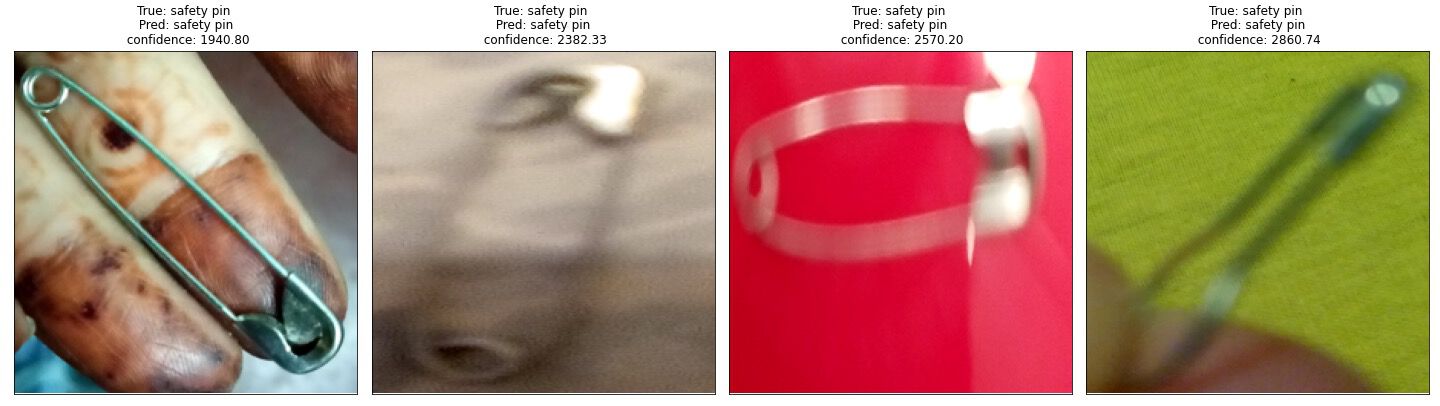}}\vspace{50pt}
\subfigure[Misclassified; highest confidences]{\includegraphics[width=1\linewidth]{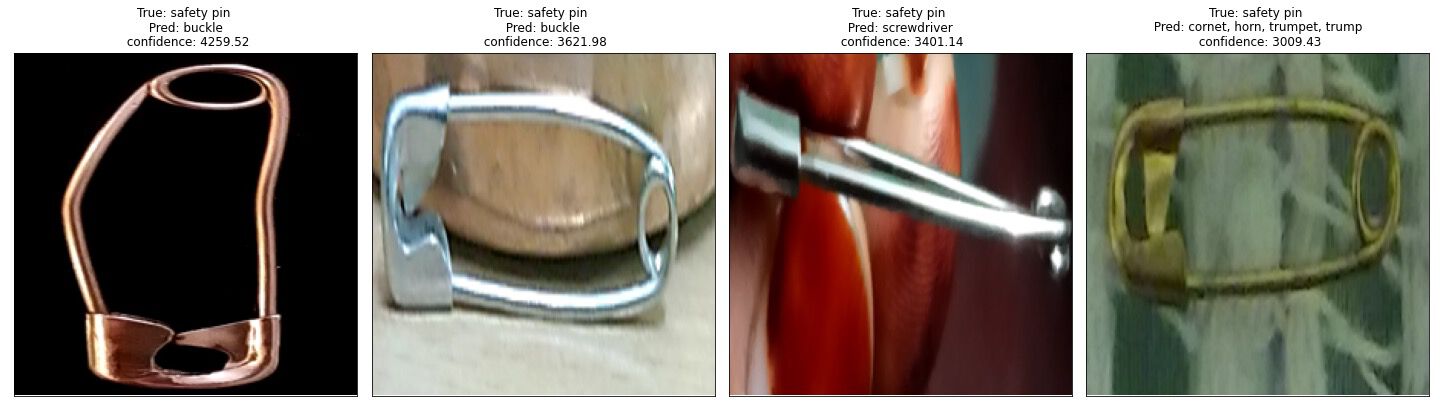}}\vspace{-10pt}
\subfigure[Misclassified; lowest confidences]{\includegraphics[width=1\linewidth]{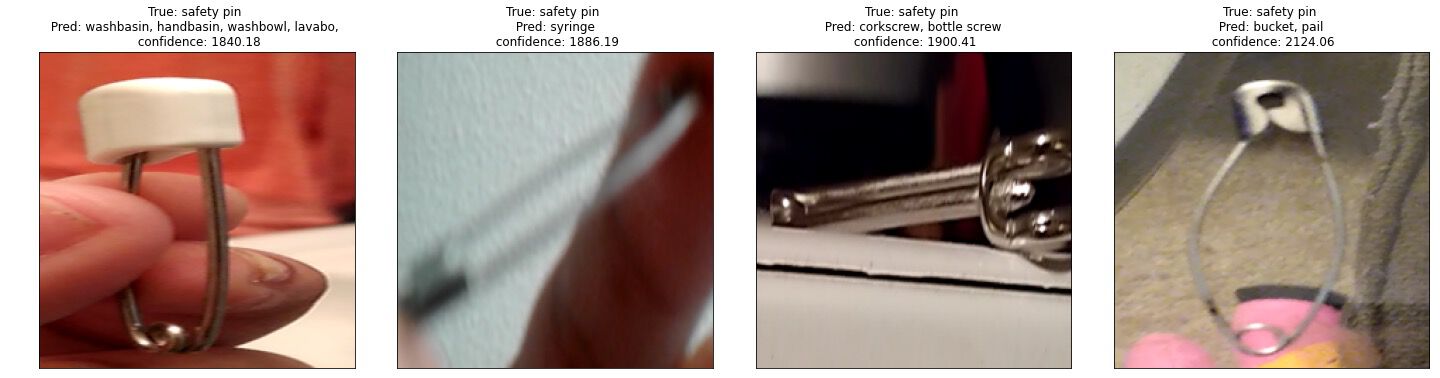}}
\caption{Correctly classified and misclassified examples from the Safety-pin class by the ResNet model.}
\label{fig:Safety-pin}
\end{figure}

\begin{figure}
\centering
\subfigure[Correctly classified; highest confidences]{\includegraphics[width=1\linewidth]{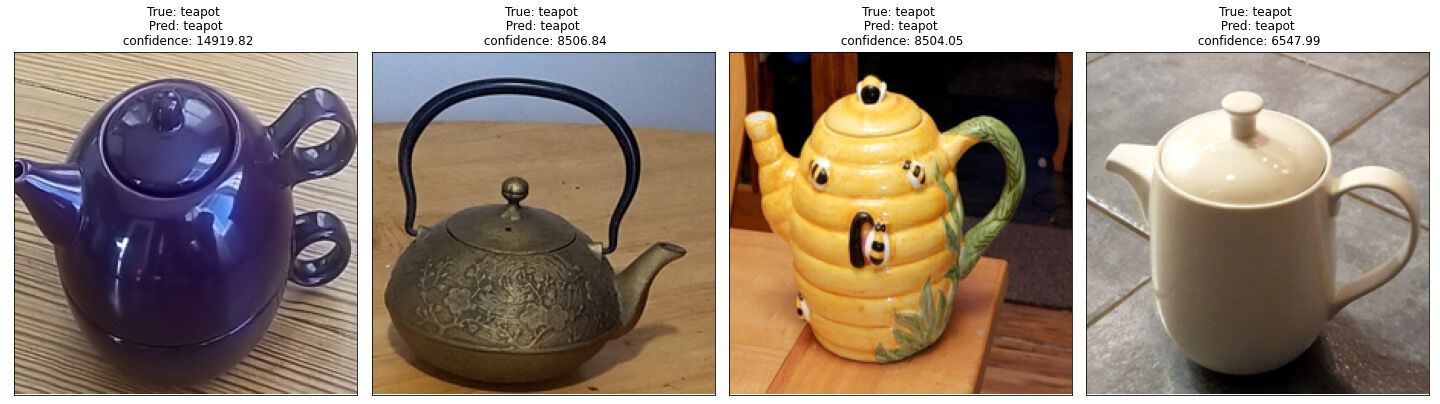}}
\vspace{-10pt}
\subfigure[Correctly classified; lowest confidences]{\includegraphics[width=1\linewidth]{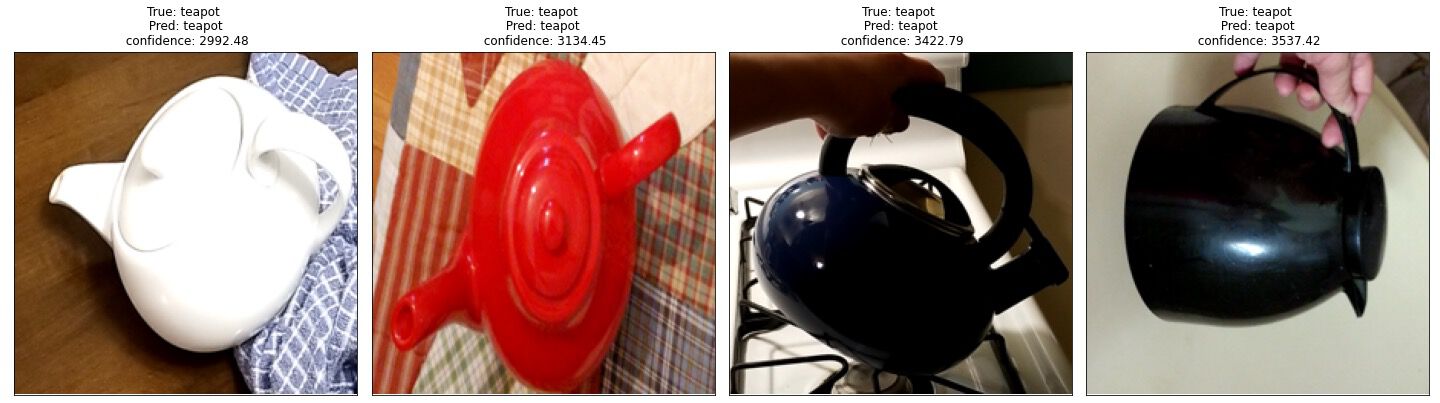}}\vspace{50pt}
\subfigure[Misclassified; highest confidences]{\includegraphics[width=1\linewidth]{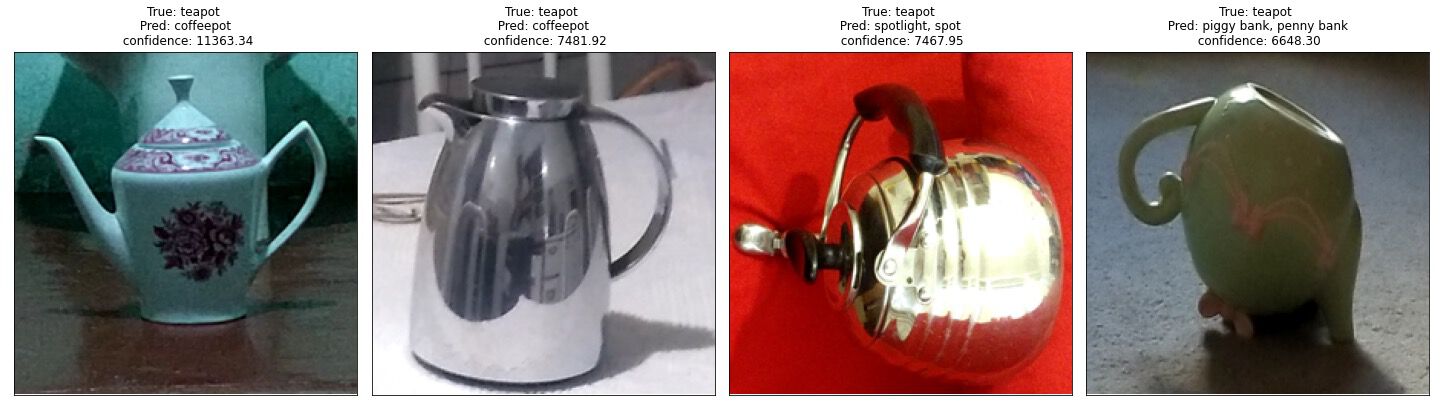}}\vspace{-10pt}
\subfigure[Misclassified; lowest confidences]{\includegraphics[width=1\linewidth]{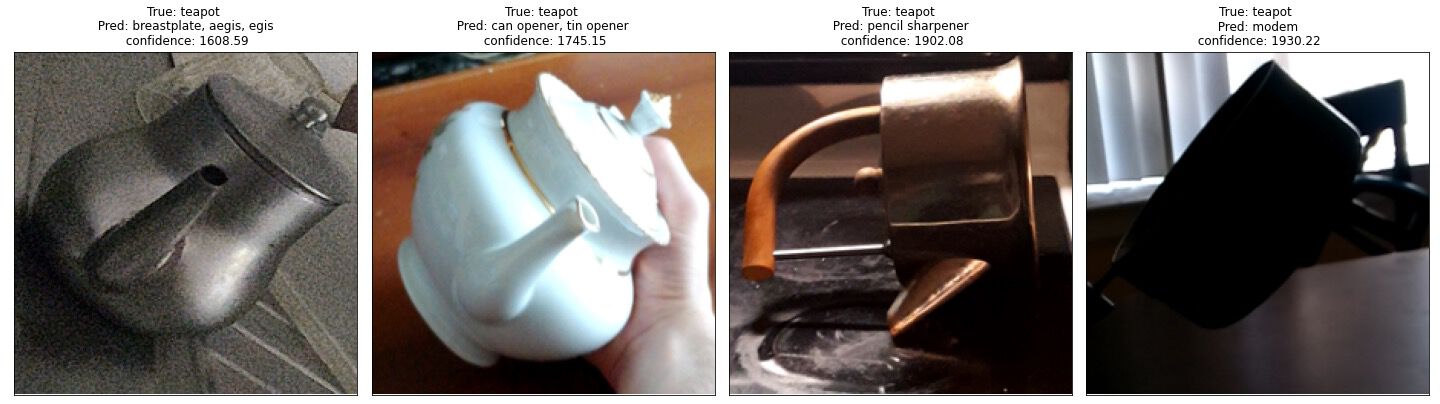}}
\caption{Correctly classified and misclassified examples from the Teapot class by the ResNet model.}
\label{fig:Teapot}
\end{figure}

\begin{figure}
\centering
\subfigure[Correctly classified; highest confidences]{\includegraphics[width=1\linewidth]{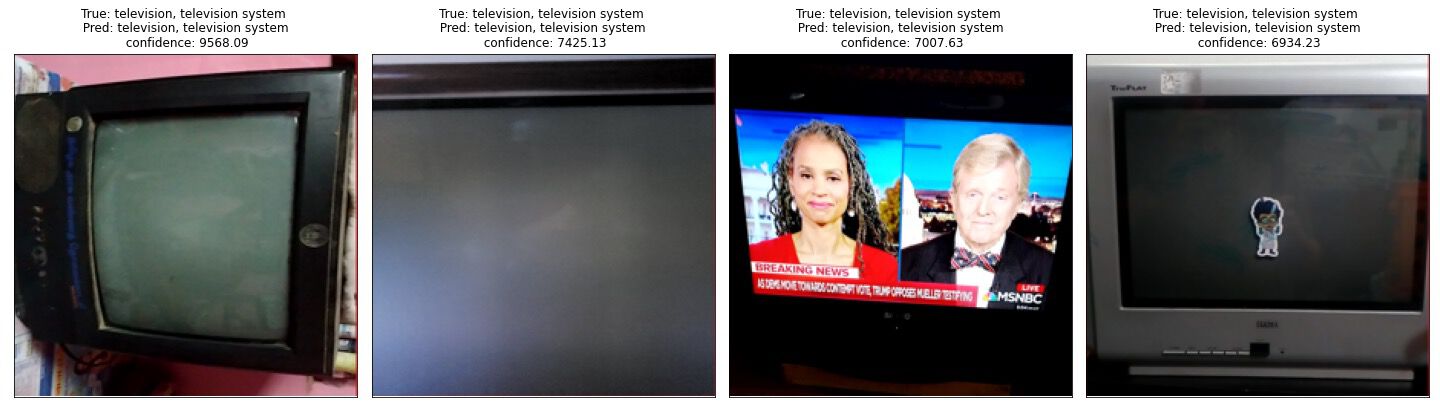}}
\vspace{-10pt}
\subfigure[Correctly classified; lowest confidences]{\includegraphics[width=1\linewidth]{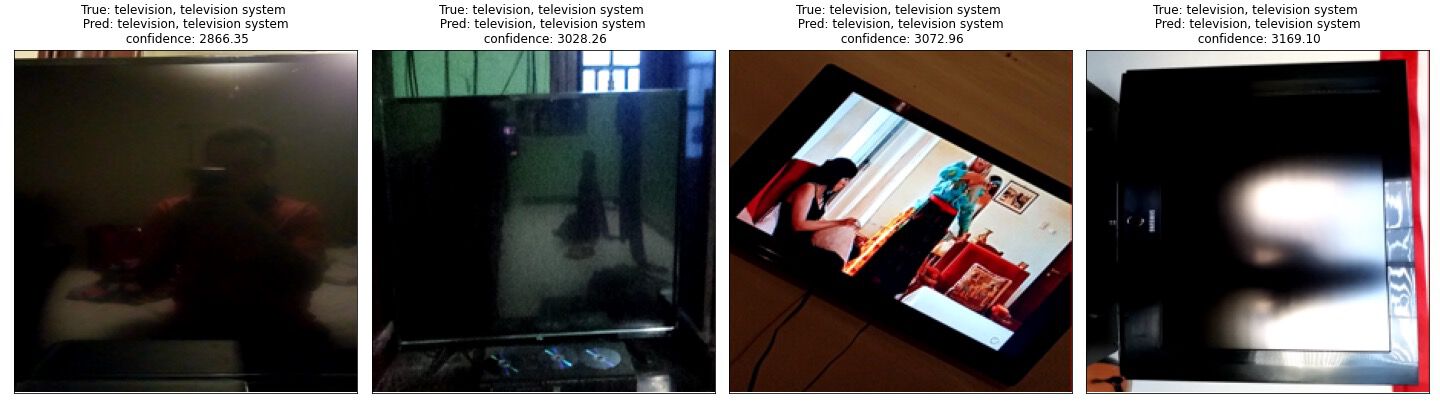}}\vspace{50pt}
\subfigure[Misclassified; highest confidences]{\includegraphics[width=1\linewidth]{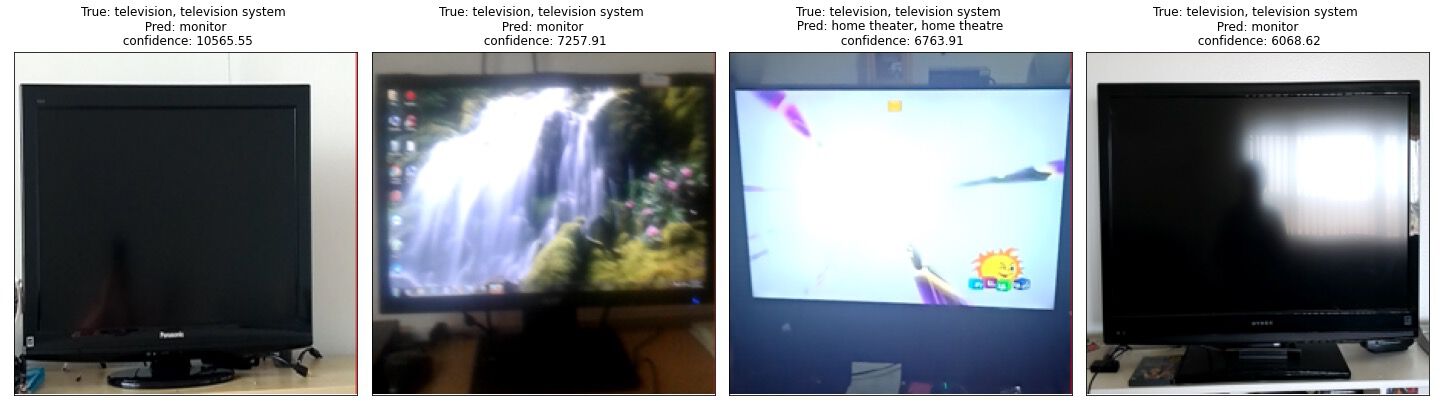}}\vspace{-10pt}
\subfigure[Misclassified; lowest confidences]{\includegraphics[width=1\linewidth]{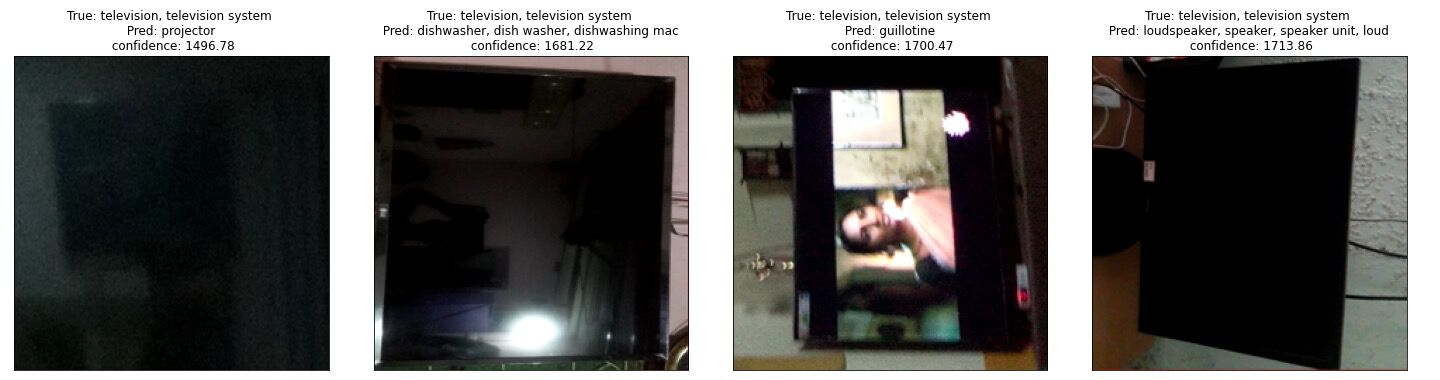}}
\caption{Correctly classified and misclassified examples from the TV class by the ResNet model.}
\label{fig:TV}
\end{figure}

\begin{figure}
\centering
\subfigure[Correctly classified; highest confidences]{\includegraphics[width=1\linewidth]{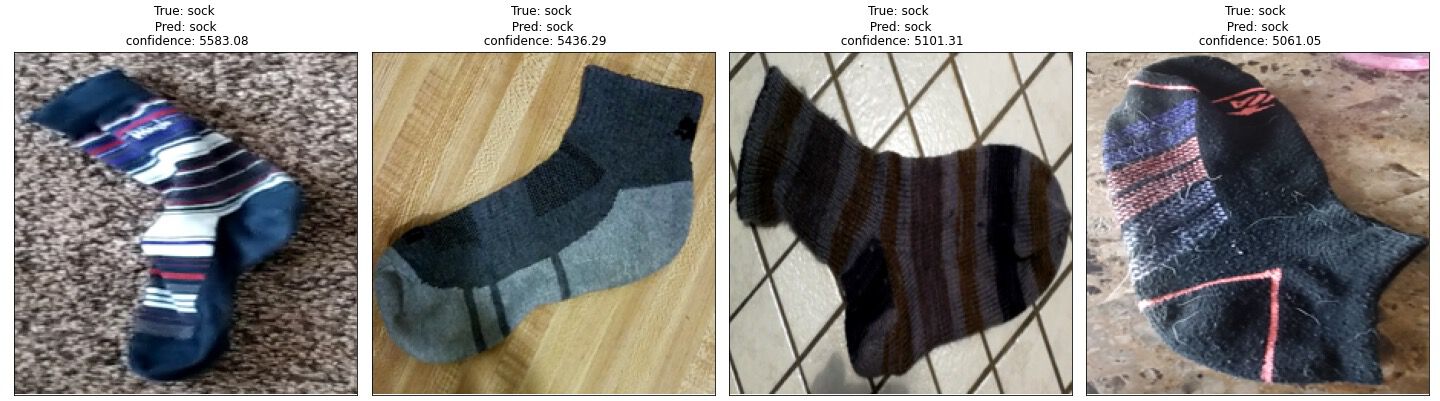}}
\vspace{-10pt}
\subfigure[Correctly classified; lowest confidences]{\includegraphics[width=1\linewidth]{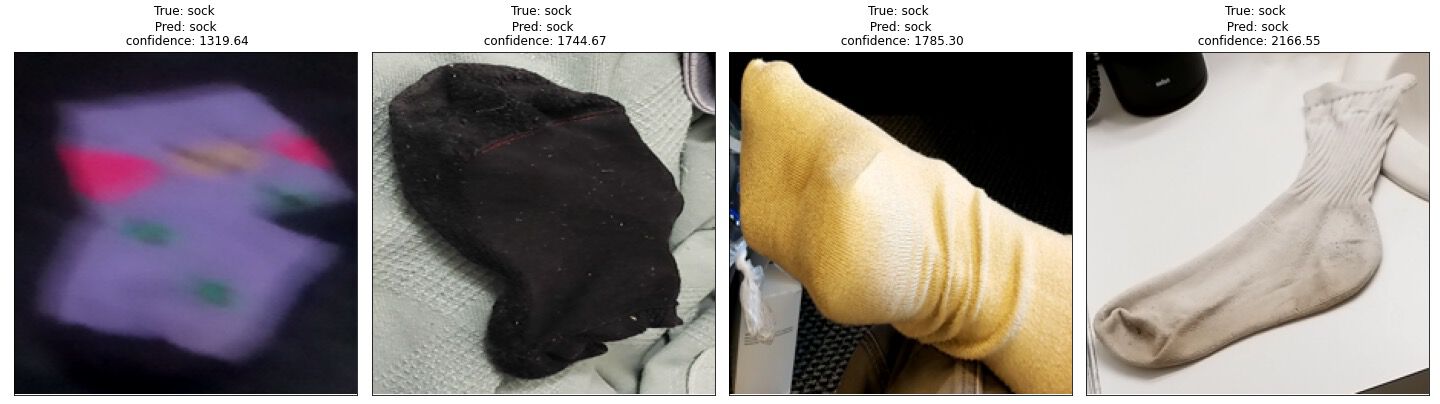}}\vspace{50pt}
\subfigure[Misclassified; highest confidences]{\includegraphics[width=1\linewidth]{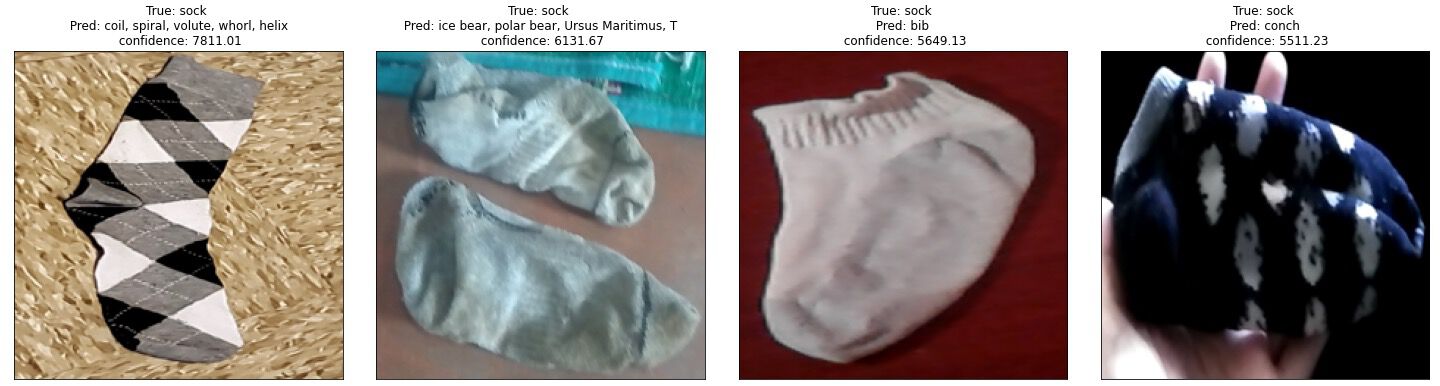}}\vspace{-10pt}
\subfigure[Misclassified; lowest confidences]{\includegraphics[width=1\linewidth]{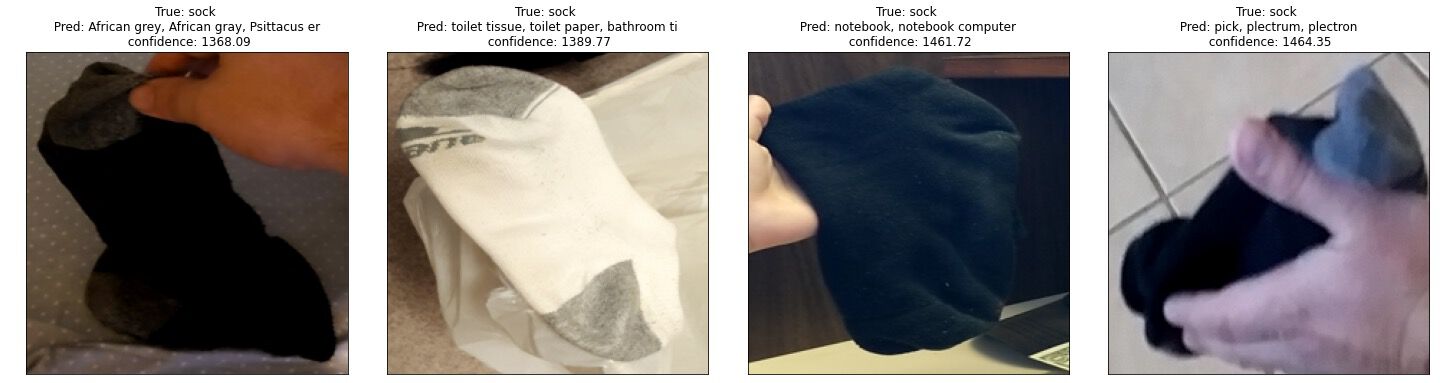}}
\caption{Correctly classified and misclassified examples from the Sock class by the ResNet model.}
\label{fig:Sock}
\end{figure}

\begin{figure}
\centering
\subfigure[Correctly classified; highest confidences]{\includegraphics[width=1\linewidth]{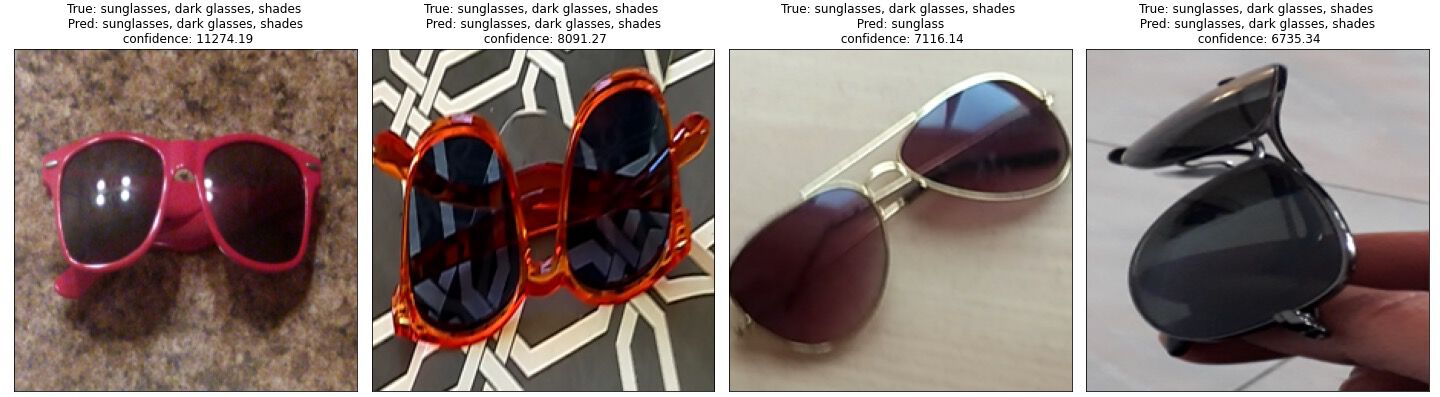}}
\vspace{-10pt}
\subfigure[Correctly classified; lowest confidences]{\includegraphics[width=1\linewidth]{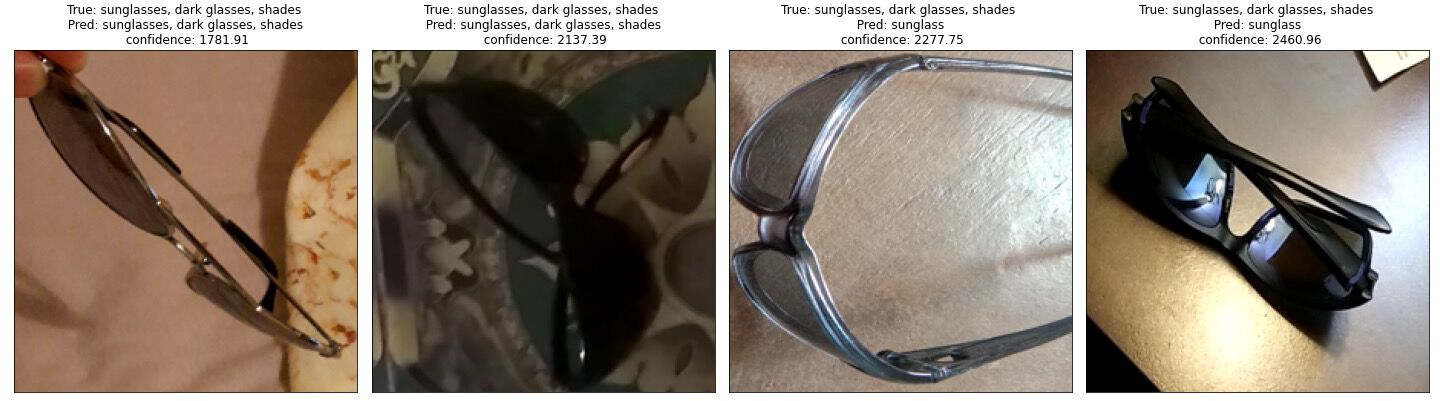}}\vspace{50pt}
\subfigure[Misclassified; highest confidences]{\includegraphics[width=1\linewidth]{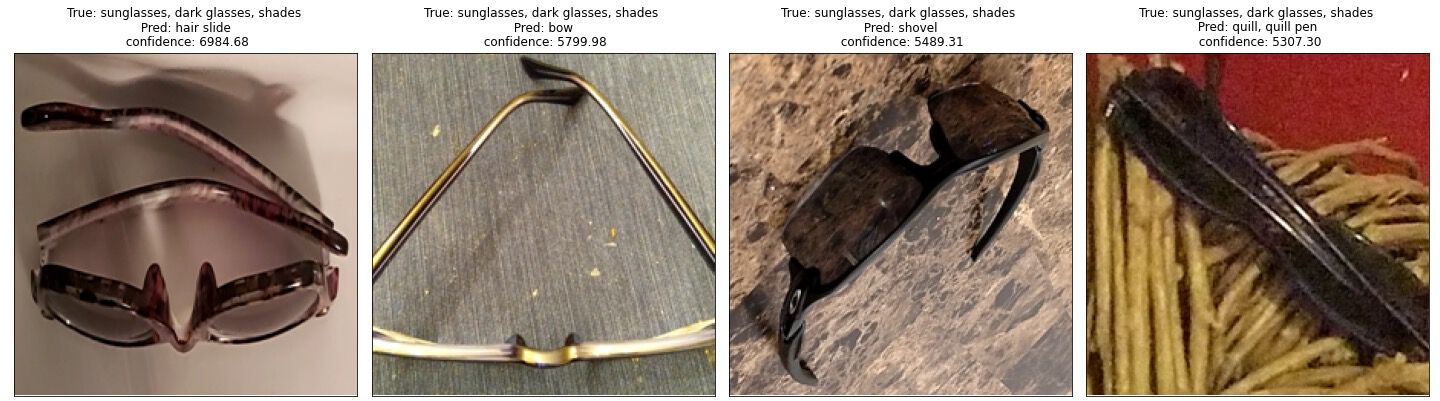}}\vspace{-10pt}
\subfigure[Misclassified; lowest confidences]{\includegraphics[width=1\linewidth]{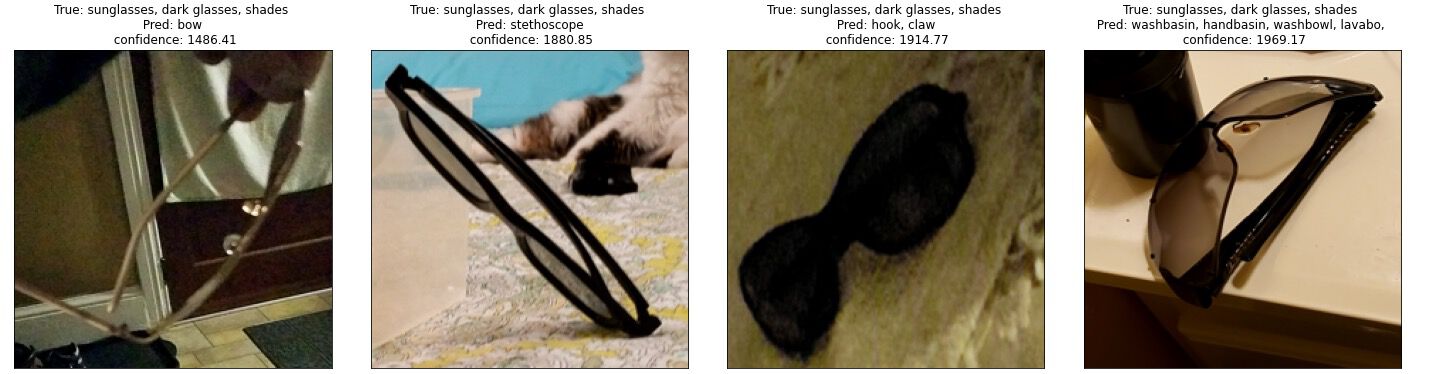}}
\caption{Correctly classified and misclassified examples from the Sunglasses class by the ResNet model.}
\label{fig:Sunglasses}
\end{figure}

\begin{figure}
\centering
\subfigure[Correctly classified; highest confidences]{\includegraphics[width=1\linewidth]{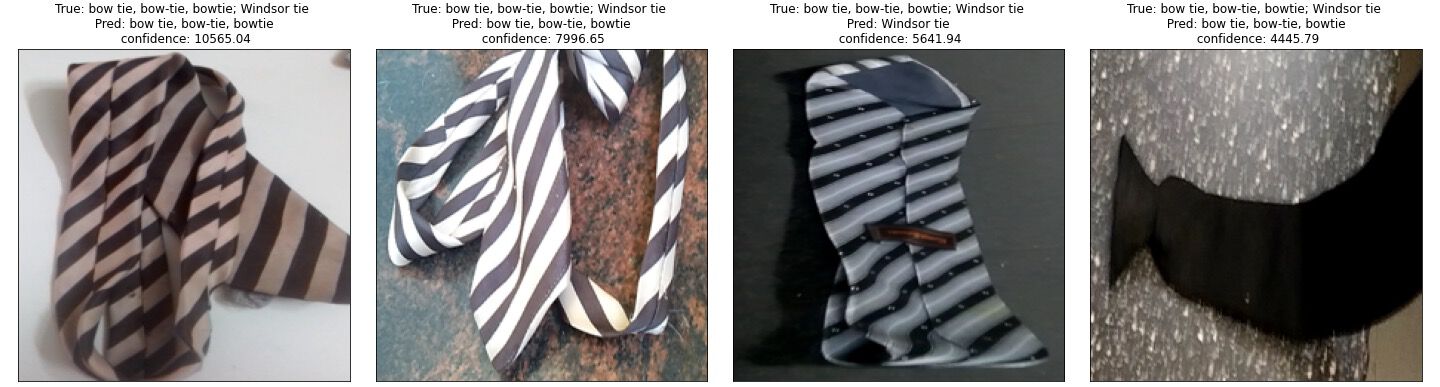}}
\vspace{-10pt}
\subfigure[Correctly classified; lowest confidences]{\includegraphics[width=1\linewidth]{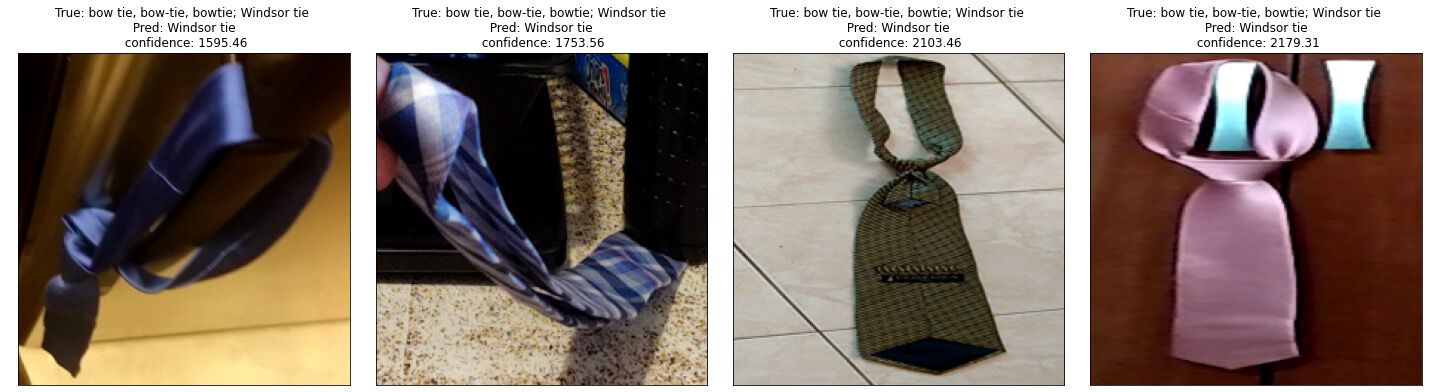}}\vspace{50pt}
\subfigure[Misclassified; highest confidences]{\includegraphics[width=1\linewidth]{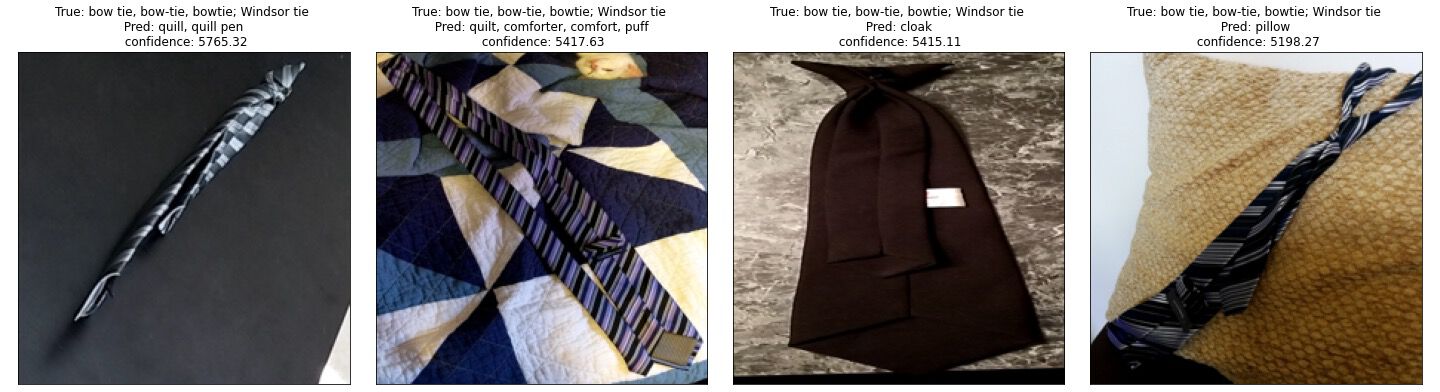}}\vspace{-10pt}
\subfigure[Misclassified; lowest confidences]{\includegraphics[width=1\linewidth]{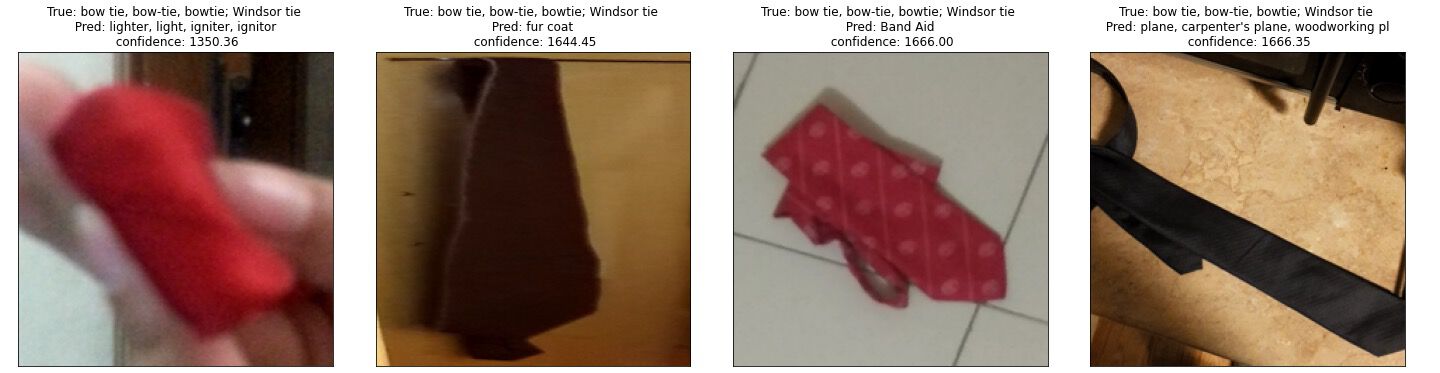}}
\caption{Correctly classified and misclassified examples from the Tie class by the ResNet model.}
\label{fig:Tie}
\end{figure}

\begin{figure}
\centering
\subfigure[Correctly classified; highest confidences]{\includegraphics[width=1\linewidth]{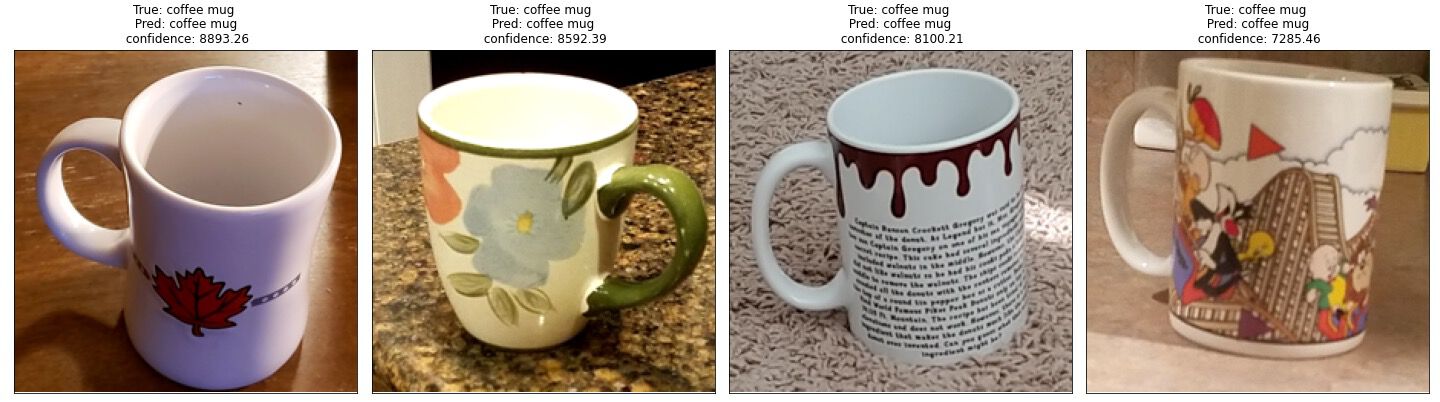}}
\vspace{-10pt}
\subfigure[Correctly classified; lowest confidences]{\includegraphics[width=1\linewidth]{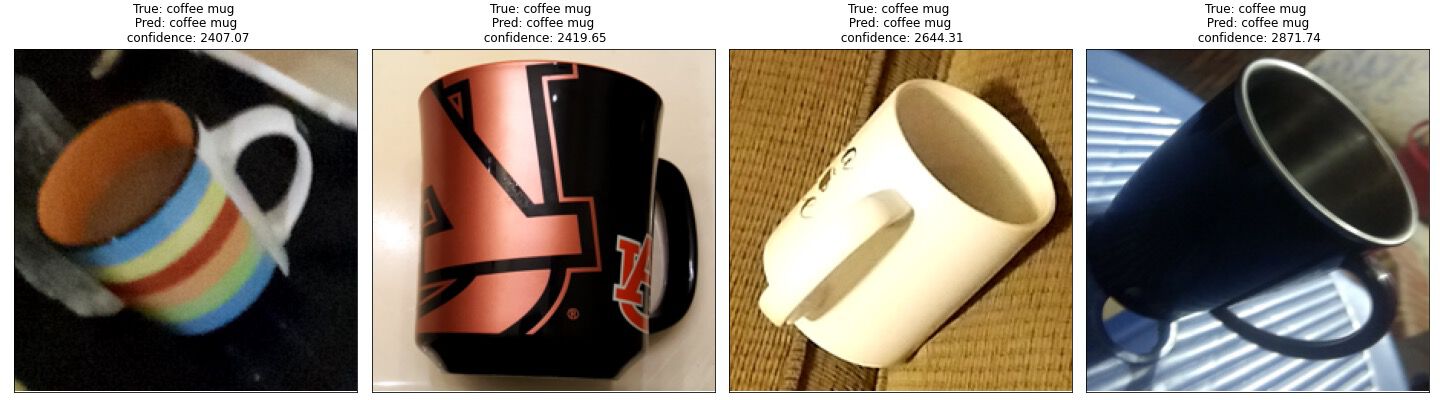}}\vspace{50pt}
\subfigure[Misclassified; highest confidences]{\includegraphics[width=1\linewidth]{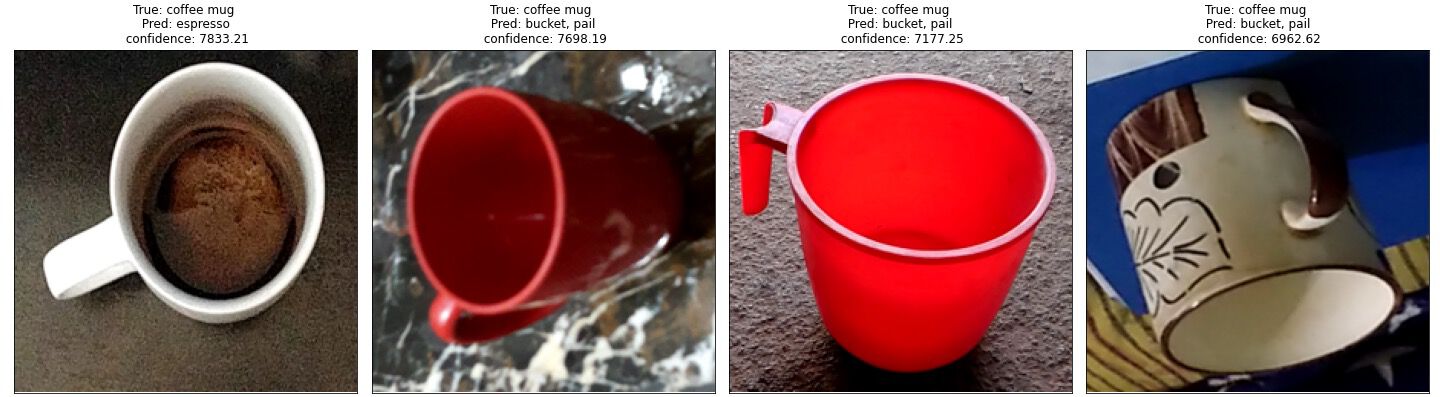}}\vspace{-10pt}
\subfigure[Misclassified; lowest confidences]{\includegraphics[width=1\linewidth]{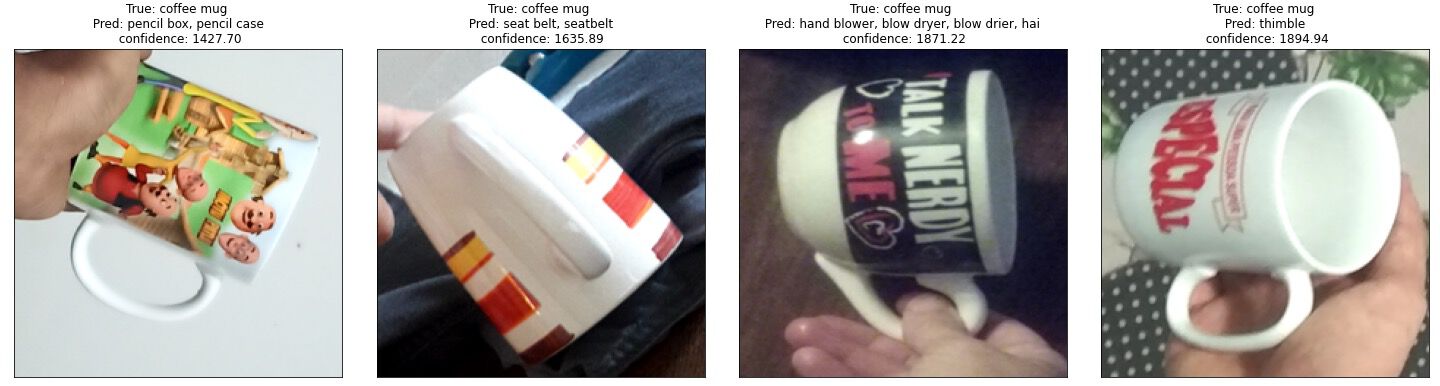}}
\caption{Correctly classified and misclassified examples from the Mug class by the ResNet model.}
\label{fig:Mug}
\end{figure}

\begin{figure}
\centering
\subfigure[Correctly classified; highest confidences]{\includegraphics[width=1\linewidth]{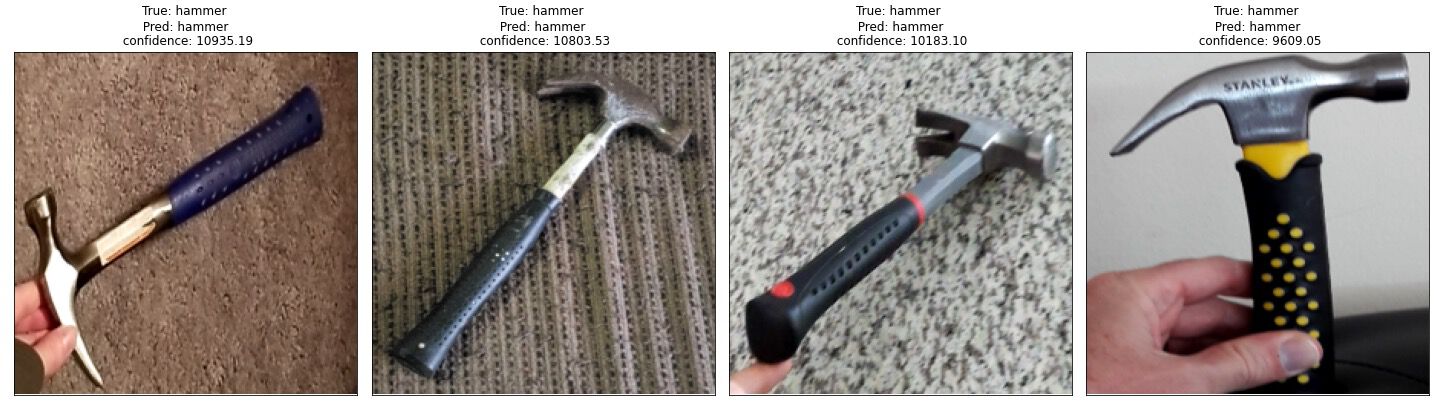}}
\vspace{-10pt}
\subfigure[Correctly classified; lowest confidences]{\includegraphics[width=1\linewidth]{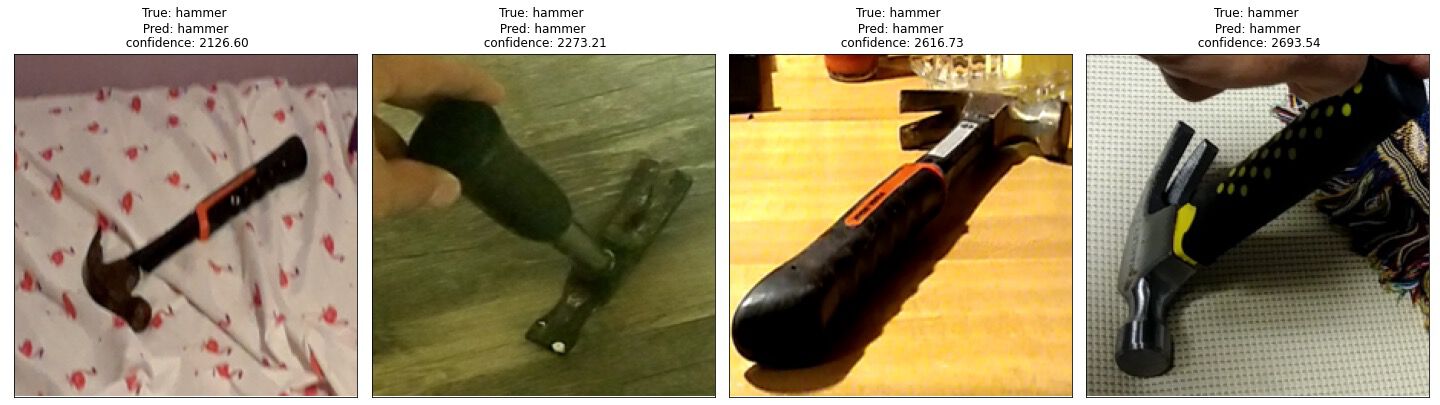}}\vspace{50pt}
\subfigure[Misclassified; highest confidences]{\includegraphics[width=1\linewidth]{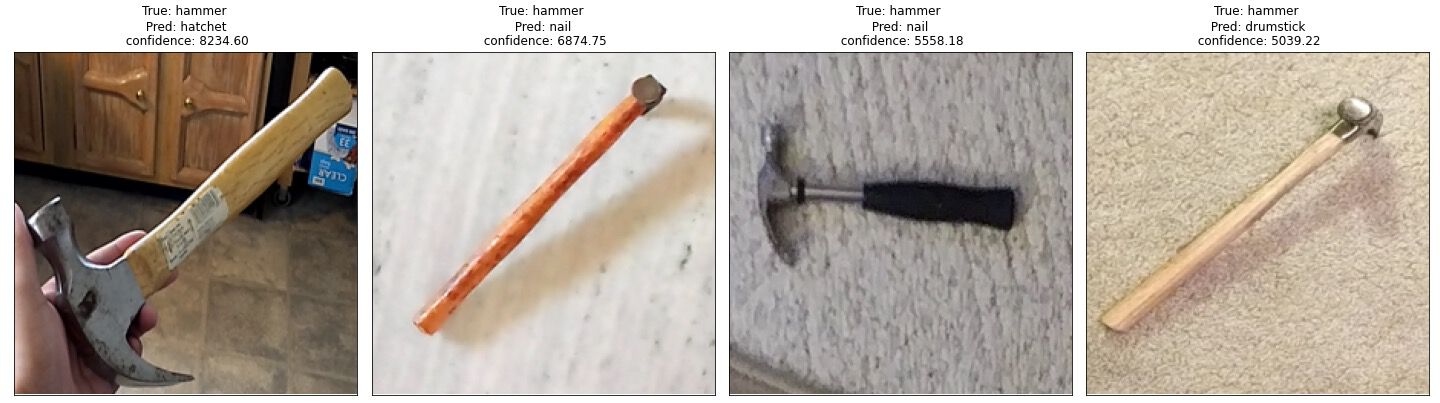}}\vspace{-10pt}
\subfigure[Misclassified; lowest confidences]{\includegraphics[width=1\linewidth]{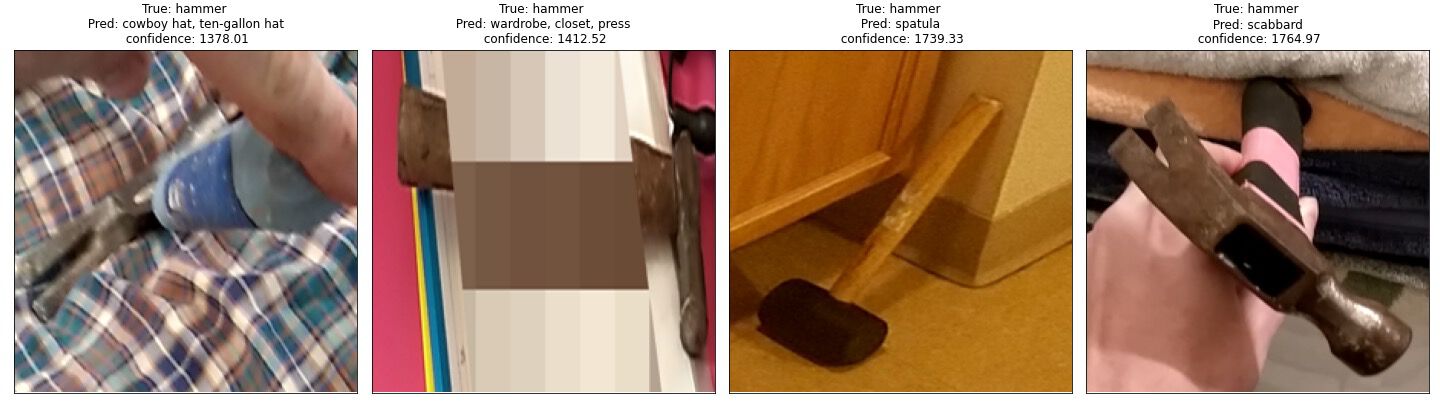}}
\caption{Correctly classified and misclassified examples from the Hammer class by the ResNet model.}
\label{fig:Hammer}
\end{figure}

\begin{figure}
\centering
\subfigure[Correctly classified; highest confidences]{\includegraphics[width=1\linewidth]{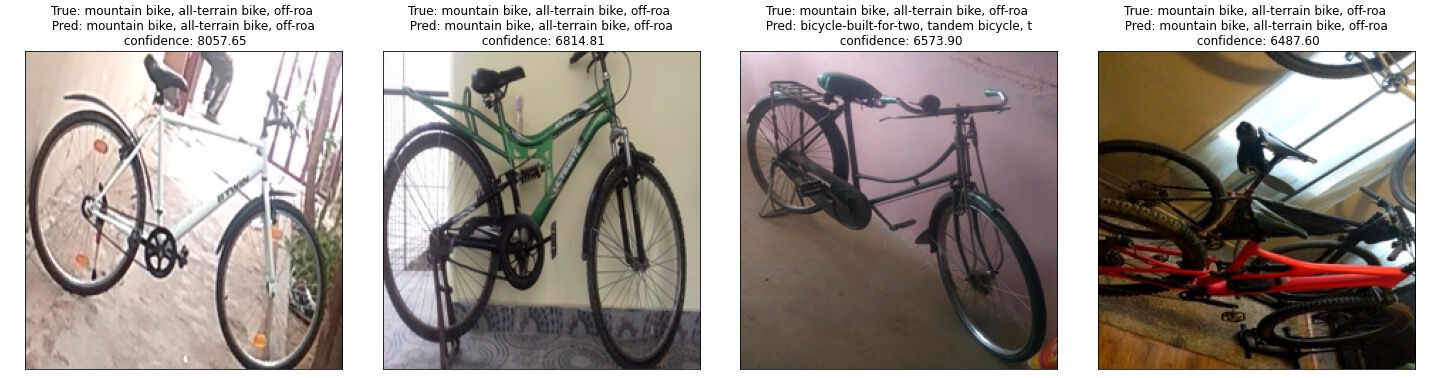}}
\vspace{-10pt}
\subfigure[Correctly classified; lowest confidences]{\includegraphics[width=1\linewidth]{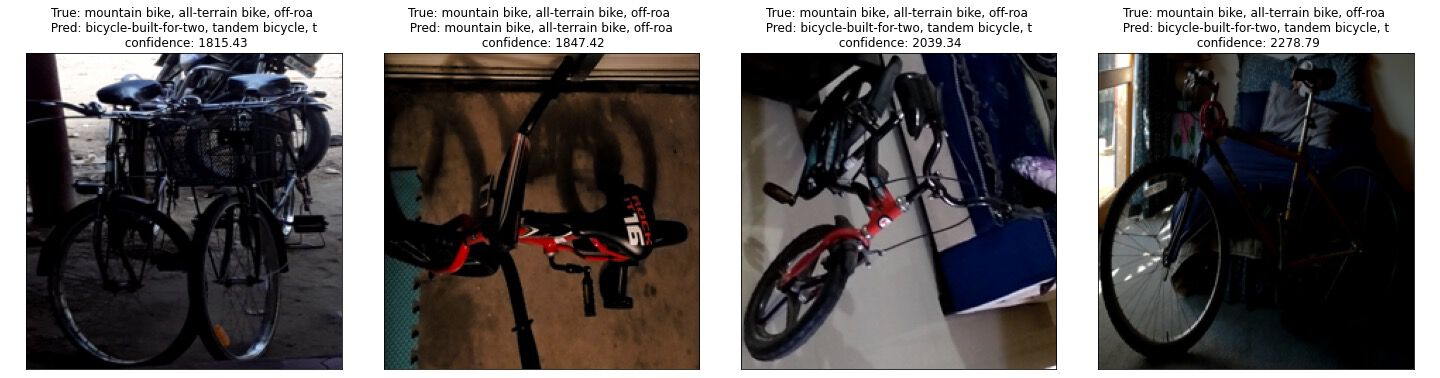}}\vspace{50pt}
\subfigure[Misclassified; highest confidences]{\includegraphics[width=1\linewidth]{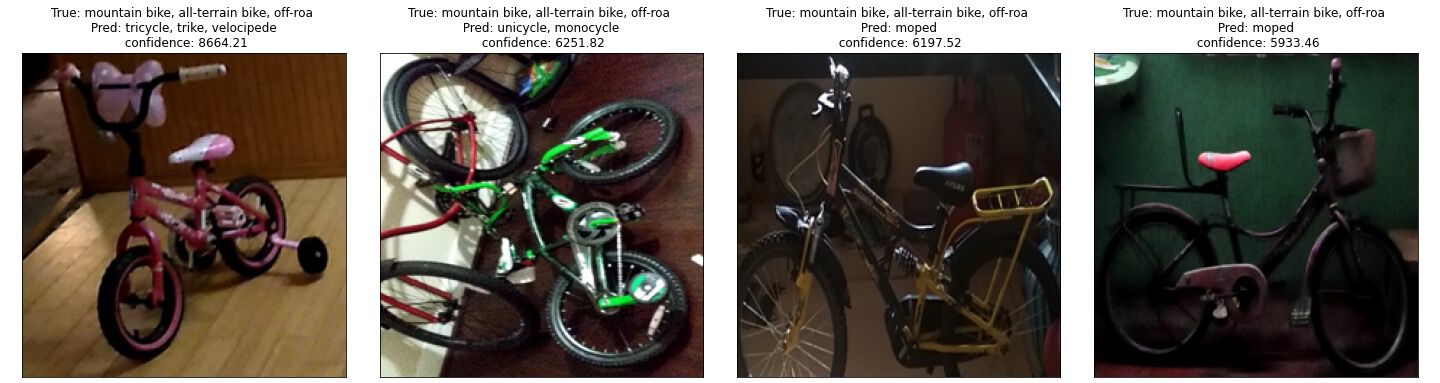}}\vspace{-10pt}
\subfigure[Misclassified; lowest confidences]{\includegraphics[width=1\linewidth]{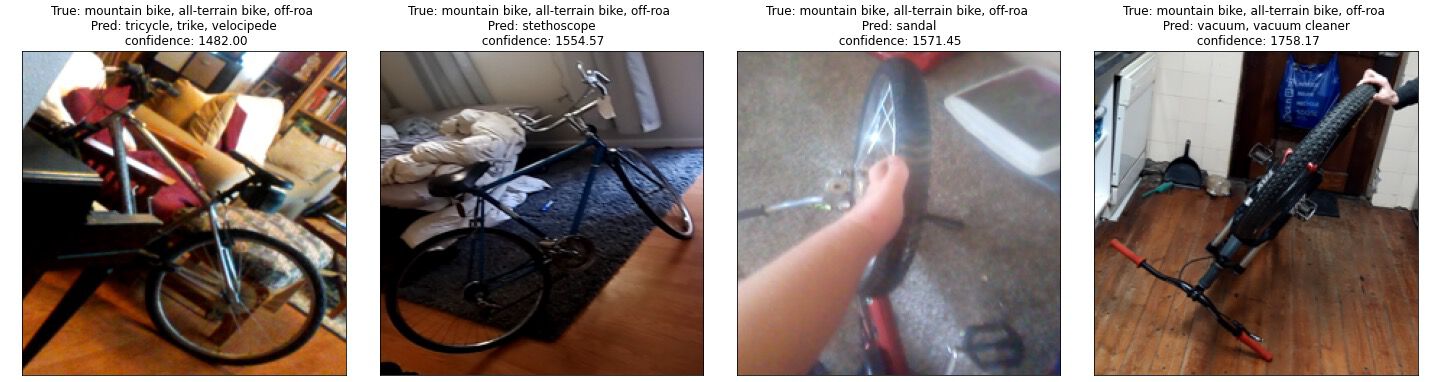}}
\caption{Correctly classified and misclassified examples from the Bicycle class by the ResNet model.}
\label{fig:Bicycle}
\end{figure}

\begin{figure}
\centering
\subfigure[Correctly classified; highest confidences]{\includegraphics[width=1\linewidth]{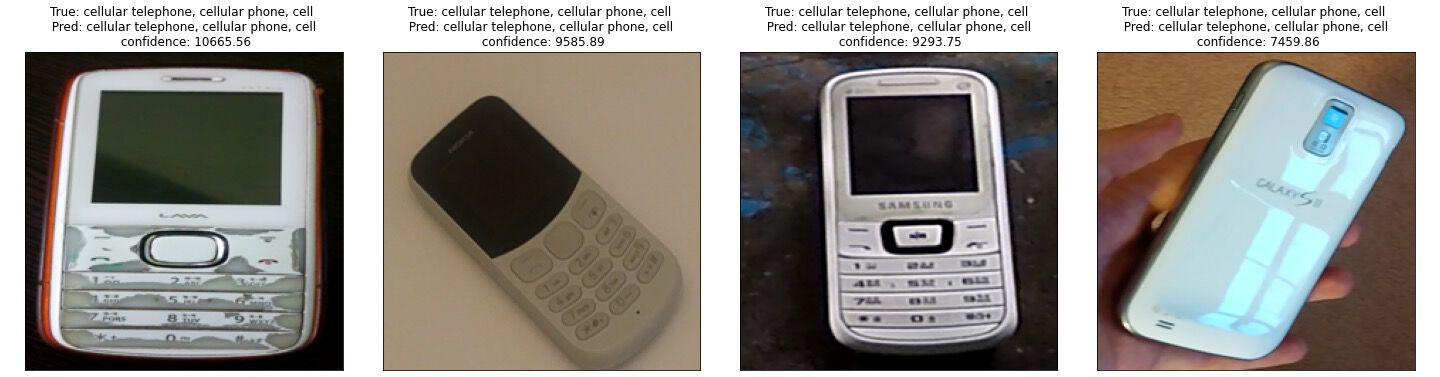}}
\vspace{-10pt}
\subfigure[Correctly classified; lowest confidences]{\includegraphics[width=1\linewidth]{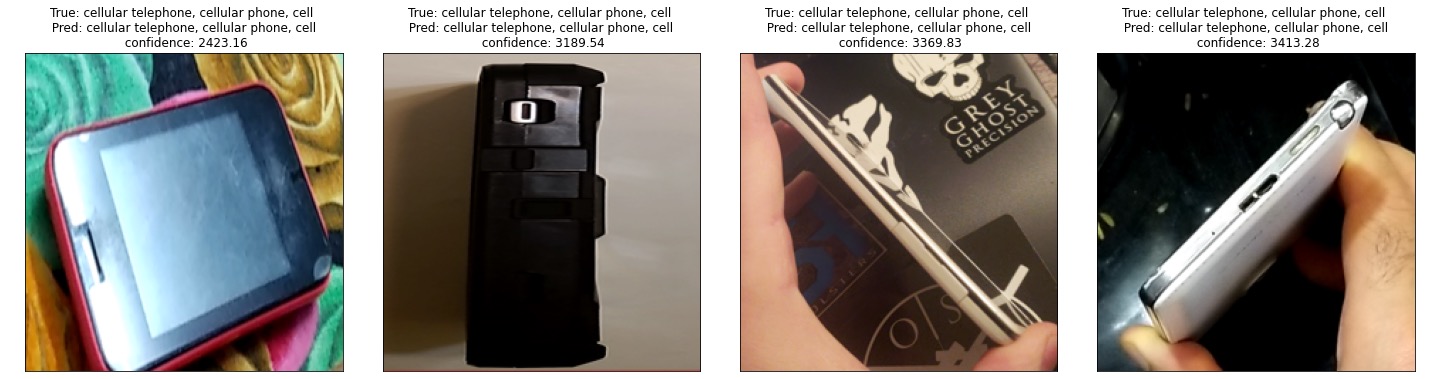}}\vspace{50pt}
\subfigure[Misclassified; highest confidences]{\includegraphics[width=1\linewidth]{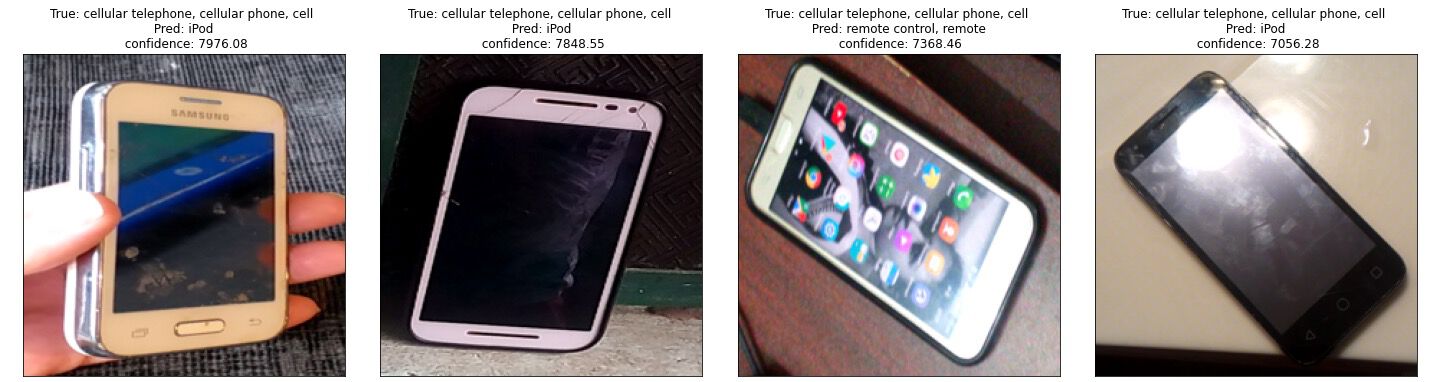}}\vspace{-10pt}
\subfigure[Misclassified; lowest confidences]{\includegraphics[width=1\linewidth]{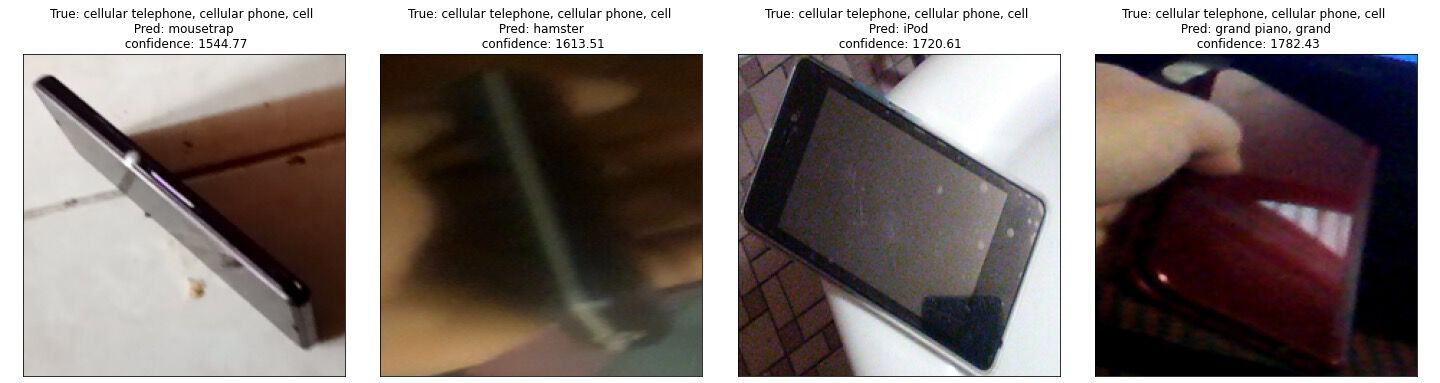}}
\caption{Correctly classified and misclassified examples from the Cellphone class by the ResNet model.}
\label{fig:Cellphone}
\end{figure}

\begin{figure}
\centering
\subfigure[Correctly classified; highest confidences]{\includegraphics[width=1\linewidth]{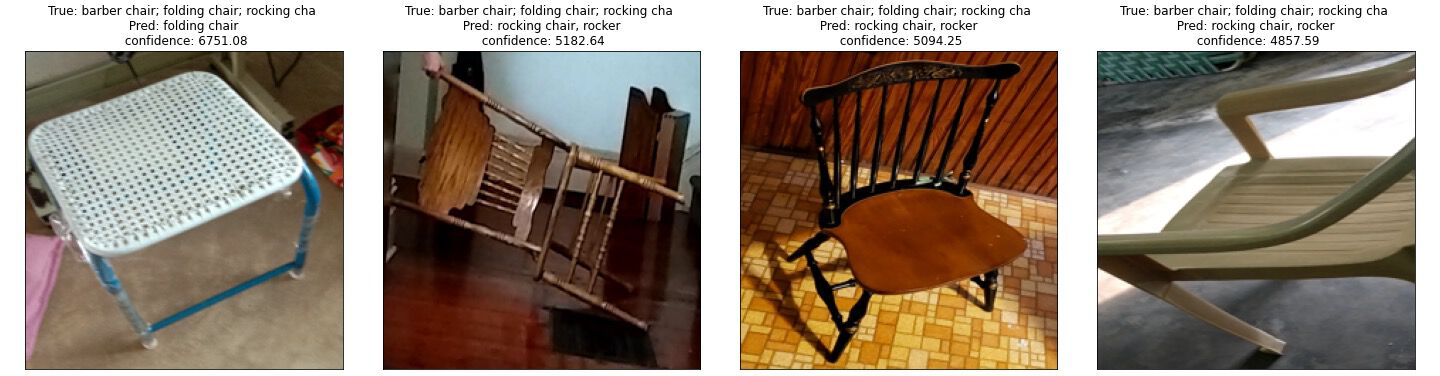}}
\vspace{-10pt}
\subfigure[Correctly classified; lowest confidences]{\includegraphics[width=1\linewidth]{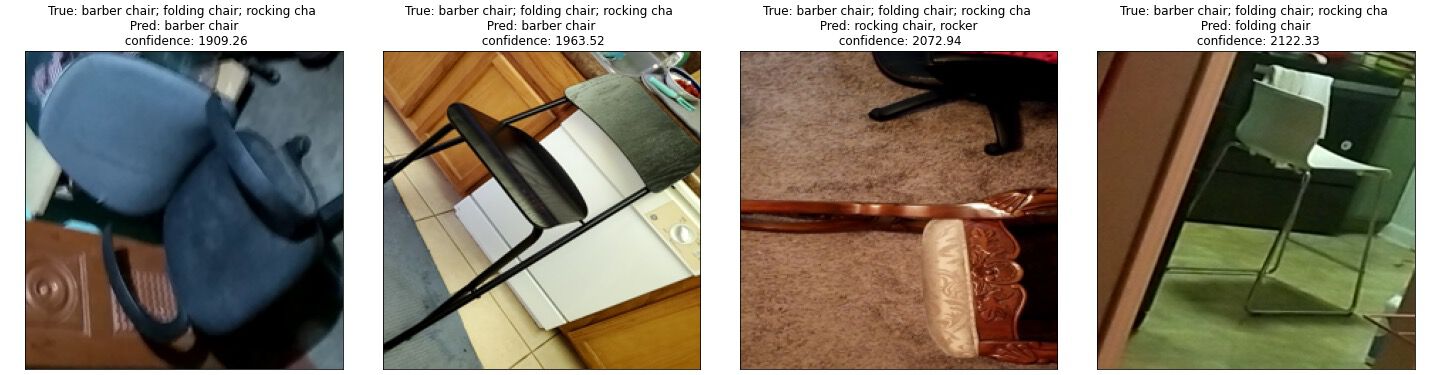}}\vspace{50pt}
\subfigure[Misclassified; highest confidences]{\includegraphics[width=1\linewidth]{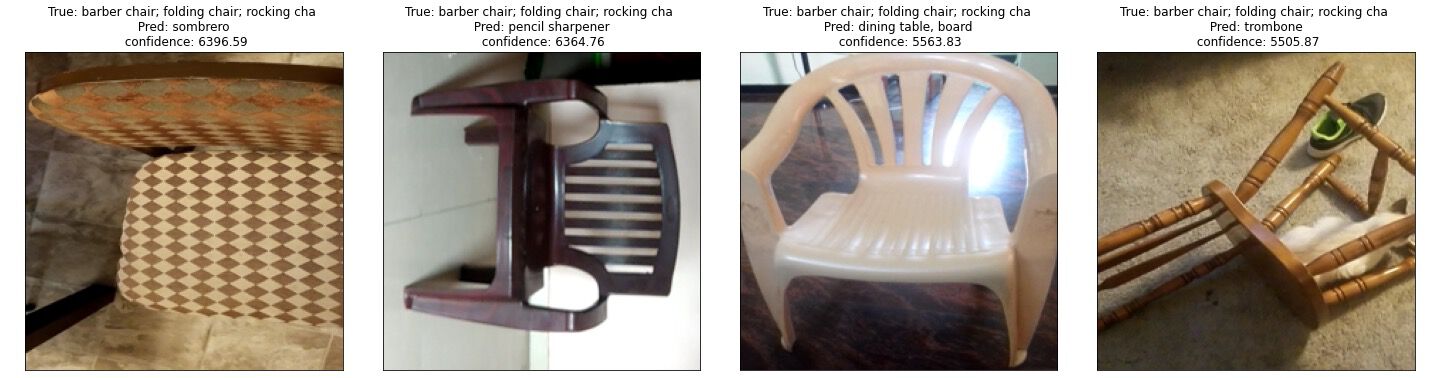}}\vspace{-10pt}
\subfigure[Misclassified; lowest confidences]{\includegraphics[width=1\linewidth]{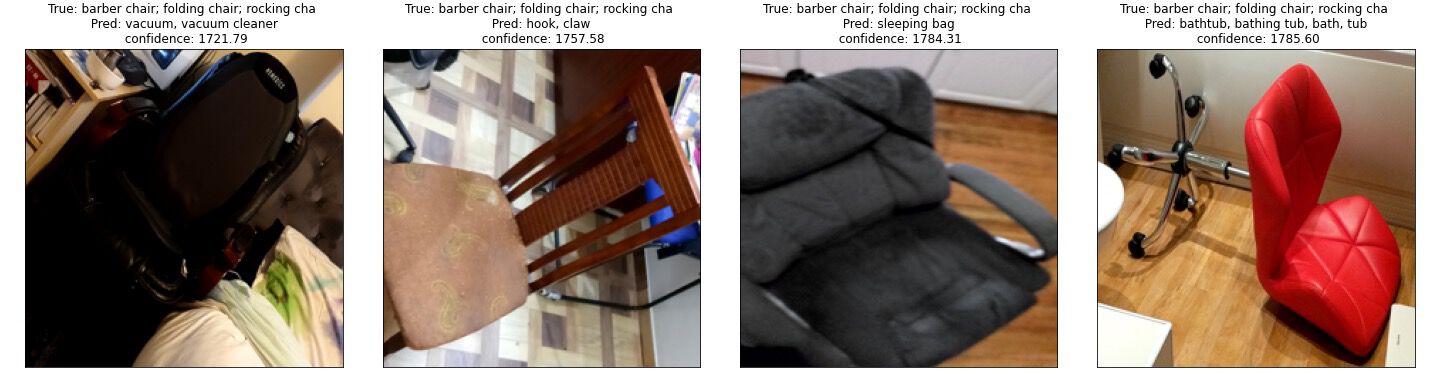}}
\caption{Correctly classified and misclassified examples from the Chair class by the ResNet model.}
\label{fig:Chair}
\end{figure}

\end{document}